\documentclass{TEAI}

\usepackage{xspace}
\usepackage{makecell}        
\usepackage{colortbl}        

\newcommand{\pfnarrow}{0.62\textwidth}
\newcommand{\pfphoto}{0.45\textwidth}   
\makeatletter
\gdef\pf@queue{}
\newcommand{\pfflush}{\pf@queue\gdef\pf@queue{}}
\newcommand{\pf@hold}[2]{\gappto\pf@queue{\@float{#1}[!htbp]}\xappto\pf@queue{\expandonce{#2}}\gappto\pf@queue{\end@float}}
\newcommand{\pfholdfloats}{%
  \RenewEnviron{figure*}[1][]{\pf@hold{figure}{\BODY}}%
  \RenewEnviron{table*}[1][]{\pf@hold{table}{\BODY}}%
  \RenewEnviron{table}[1][]{\pf@hold{table}{\BODY}}}
\let\pf@origtable\table \let\pf@origendtable\endtable
\newcommand{\pfrestorefloats}{\pfflush\let\table\pf@origtable\let\endtable\pf@origendtable}
\makeatother

\usepackage{environ}
\let\pfsetabstract\abstract
\RenewEnviron{abstract}{%
  \expandafter\pfsetabstract\expandafter{\BODY}%
  \global\let\abstractlist\abstractlist}
\NewEnviron{IEEEkeywords}{}   

\newcommand{\method}{ProgressFlip\xspace}          
\newcommand{\qdwbc}{QDW-BC\xspace}                 
\newcommand{\qdwbcm}{\mathrm{QDW\text{-}BC}}       

\definecolor{pfcell}{HTML}{4472C4}

\title{\method: Learning Frame-Level Relative Progress for Long-Horizon Tasks
from Human Interventions}

\author[1,2*]{Qiuyi Gu}
\author[1,2*]{Zhipeng Hu}
\author[3]{Jinyang Xiao}
\author[2\dagger]{Feng Gao}
\author[1,2]{Shu'ang Yu}
\author[5]{Zhihao Liu}
\author[1,2]{Wenhao Tang}
\author[2]{Pengcheng Xia}
\author[2]{Yuanfan Xu}
\author[3,4]{Xiaodan Liang}
\author[1]{Jincheng Yu}
\author[1]{Yu Wang}
\author[1\ddagger]{Chao Yu}
\author[1\ddagger]{Xinlei Chen}
\affiliation[1]{Tsinghua University}
\affiliation[2]{Striding AI}
\affiliation[3]{Sun Yat-sen University}
\affiliation[4]{Pengcheng Laboratory}
\affiliation[5]{Institute of Automation, Chinese Academy of Sciences}
\contribution[*]{Equal contribution}
\contribution[\dagger]{Project leader}
\contribution[\ddagger]{Equal advising}

\renewcommand{\pfprojecturl}{https://progressflip.github.io}
\date{November 28, 2026}   

\begin{document}

\maketitle

\begin{abstract}
Post-training of robot foundation models increasingly relies on mixed-quality interaction data, which interleaves autonomous rollouts, failure-prone behaviors, and manual recoveries within single trajectories. 
Exploiting such data requires fine-grained credit assignment, yet accurately evaluating real failure states remains challenging, especially for long-horizon tasks where failures can emerge at any subtask stage. 
We observe a key phenomenon in human intervention data where task progress degrades as the policy enters a failure and resumes advancement once human recovery begins. 
Building on this insight, we present \method, a progress reward modeling framework that quantifies relative progress through signed temporal offsets between observation pairs. 
Trained on both expert demonstrations and human intervention data, \method provides scalable, frame-level supervision on real physical failures and recoveries without manual annotation. 
We further introduce Quality-Difficulty Weighted Behavior Cloning (\qdwbc) to resolve stage-level budget imbalance by weighting policy updates with both frame progress and stage difficulty. 
Across four RoboDojo simulation tasks and two multi-stage real-world tasks, our method improves the average success rate by $11.6\%$ over the state-of-the-art baseline. 
Specifically, it achieves $90.0\%$ success on table clearing and $76.7\%$ on whiskey sour making, compared with $6.7\%$ and $0\%$ for vanilla BC.
\end{abstract}

\teaiabstract

%
\begin{figure}[!b]
  \centering
  \includegraphics[width=0.95\textwidth]{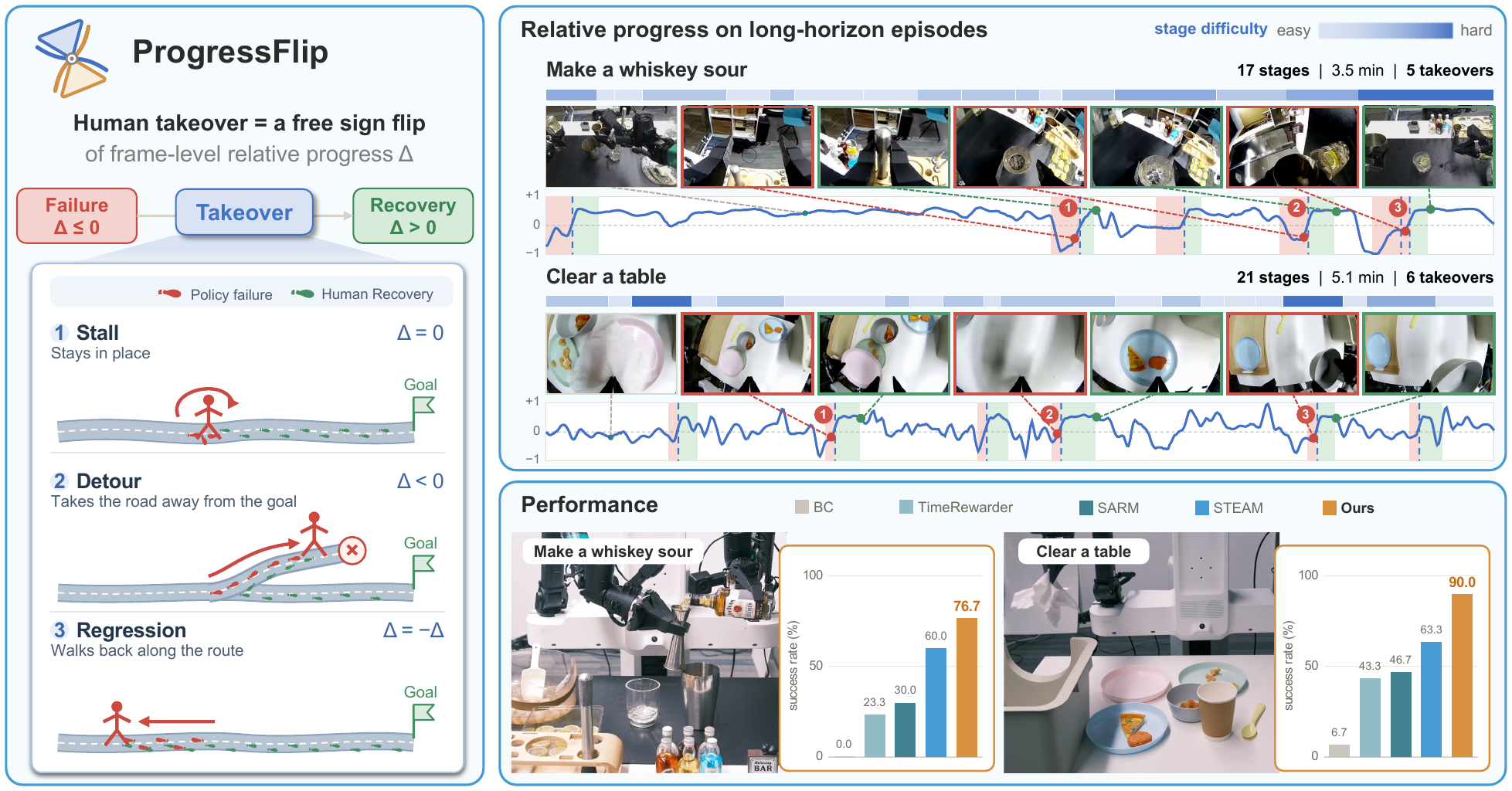}
  \caption{\textbf{Overview.}
  \textbf{Left.} Policy failures encompass stalling, detouring, and regressing. Human takeovers mark a progress sign flip where relative progress transitions from non-positive failure to positive recovery.
  \textbf{Top right.} Predicted relative progress on long-horizon episodes, where top color bars indicate stage difficulty. Red and green shading denote pre-takeover failure segments and post-takeover recovery segments, with numbered markers locating intervention events.
  \textbf{Bottom right.} Downstream policy success rates.}
  \label{fig:overview}
\end{figure}

\pfholdfloats
\section{Introduction}
\label{sec:intro}

While large-scale pretraining endows robot foundation models with general manipulation capabilities \cite{black2024pi_0,intelligence2025pi_,ye2026world}, deploying them effectively on complex, long-horizon tasks requires post-training paradigms that enable policies to learn from experience \cite{intelligence2025pi}. 
A primary bottleneck in robotic post-training lies in equipping policies with the capability to recover from execution errors, where human-in-the-loop intervention \cite{dong2026expo, hu2025rac, mao2026arm} has emerged as a data-efficient approach. 
By allowing policies to explore autonomously and requiring human operators to intervene only during imminent failures, human effort is strategically concentrated on the most informative states. 
However, this interactive process naturally yields highly heterogeneous, mixed-quality datasets that blend autonomous rollouts, failure-prone behaviors, and manual recoveries. 
Effectively leveraging such mixed-quality data for policy improvement demands a fine-grained, frame-level credit assignment framework.

Progress Reward Models (PRMs) \cite{pmlr-v305-zhang25a, chen2026sarm, liu2026steam, liang2026robometer, torne2026freeform, intelligence2025pi, yang2026rise, zhai2026help, mao2026arm} offer a promising solution by evaluating execution quality within mixed-quality interactions. 
While effective for short-horizon tasks, adapting PRMs to long-horizon manipulation remains challenging due to the difficulty of obtaining scalable, frame-level supervision on real physical failures. 
Existing approaches attempt to address this challenge through three primary paradigms. 
The first category trains exclusively on expert demonstrations, constructing synthetic negatives through heuristic techniques such as trajectory rewinding, counterfactual text, or temporal truncation \cite{pmlr-v305-zhang25a, chen2026sarm, liu2026steam}. 
However, lacking exposure to physical failures during training, these models systematically overestimate progress scores when encountering real failures. 
A second category leverages interaction data with sparse, binary success indicators, either through pairwise trajectory comparisons for preference learning \cite{liang2026robometer, torne2026freeform}, or by propagating terminal outcomes backward \cite{intelligence2025pi, yang2026rise}. 
Nonetheless, these sparse objectives struggle to provide accurate frame-level evaluations, as critical failures can occur at any stage throughout long-horizon execution. 
The third category obtains dense, frame-level supervision by directly annotating trajectories frame by frame \cite{zhai2026help, mao2026arm}, which is prohibitively expensive to scale due to continuous human labeling.

To address these limitations, we introduce \method, a progress reward modeling framework that quantifies relative progress through signed temporal offsets between observation pairs. 
Our key insight is that a human intervention provides a free annotation of a progress sign flip. 
Frames immediately preceding a takeover capture an unrecoverable failure, driving relative progress into negative values, while subsequent frames represent human recovery, restoring progress to positive values. 
By transforming takeover events into sign-flip supervision, \method provides a scalable, frame-level training signal on real physical failures and recoveries, directly addressing the overestimation and scalability bottlenecks of prior progress reward models.

Long-horizon tasks consist of multiple subtask stages that vary widely in difficulty. 
Beyond frame-level quality signals, the frequency and distribution of human takeovers across subtasks naturally reveal the specific bottleneck stages where the policy is most prone to fail. 
Vanilla behavior cloning (BC) and its weighted variants \cite{peng2019advantage, nair2020awac} allocate learning capacity proportional to stage duration rather than stage difficulty, leaving brief but critical bottleneck phases under-optimized. 
To resolve this imbalance, we design a difficulty-aware policy extraction strategy that estimates stage-wise conditional failure rates from takeover statistics to rebalance policy capacity toward challenging subtasks.

In summary, our primary contributions are three-fold.
\begin{itemize}
\item We present \method, a progress reward modeling framework that learns relative progress from expert demonstrations and human interventions. By partitioning subtask stages into sign-supervised failure segments and offset-supervised recovery segments, \method provides faithful predictions on real physical failures and recoveries.
\item We introduce Quality-Difficulty Weighted Behavior Cloning, a policy extraction strategy that resolves stage-level budget imbalances in long-horizon tasks. By estimating conditional failure rates from stage-wise takeover frequencies, \qdwbc combines frame progress with stage difficulty to focus policy capacity on challenging subtasks.
\item We validate our post-training approach across four RoboDojo simulation tasks and two multi-stage real-world tasks, including table clearing and whiskey sour making. Our approach outperforms STEAM, the strongest baseline, by an average of 11.6\% and achieves $90.0\%$ and $76.7\%$ success rates on the two real-world tasks, respectively.
\end{itemize}

\section{Related Work}
\label{sec:related}

%
\subsection{Progress Reward Models for Manipulation}
\label{sec:related:value}
Progress Reward Models evaluate execution progress toward task completion, converting sparse episode outcomes into dense frame-level signals used for reinforcement learning rewards \cite{pmlr-v306-liu26et}, advantage estimation in policy optimization \cite{chen2026sarm, mao2026arm}, or data filtering \cite{liang2026robometer}.
Most approaches \cite{liang2026robometer, pmlr-v305-zhang25a, huang2026rynnvalue} predict absolute progress by assuming monotonic task advancement and defining expert progress via normalized time indices.
However, this absolute formulation fails in long-horizon manipulation, where temporal duration diverges from semantic progress due to varying action difficulties and execution speeds \cite{chen2026sarm}.
To mitigate this, stage-aware methods partition long-horizon tasks into subtasks and estimate local progress within each individual stage \cite{chen2026sarm, kim2025reds, hung2025victor}.

Alternatively, relative progress methods predict signed temporal offsets \cite{liu2026steam, pmlr-v306-liu26et, zhanggeneralist} or discrete progress states \cite{mao2026arm}, bypassing global temporal anchors. 
Beyond manual annotations \cite{mao2026arm}, self-supervised approaches synthesize negatives by reversing expert videos \cite{liu2026steam, zhanggeneralist}. 
However, such synthetic regression fails to capture real physical failures, leading to systematic overestimation on out-of-distribution states \cite{qin2026retvl}. 
We leverage human interventions as a free supervisory signal for physical failures and recoveries, resolving overestimation without relying on synthetic data or manual annotations.

\subsection{Policy Extraction from Mixed-quality Data}
\label{sec:related:learn}
Post-training of robot foundation models increasingly relies on human-in-the-loop interactions, yielding mixed-quality datasets that combine progress, stalls, failures, and recoveries \cite{hu2025rac, intelligence2025pi, xiao2026rove, liu2026steam, yang2026aloe}.
Algorithms leveraging such data fall into two main categories.
Interactive imitation learning (DAgger-style) expands state coverage via expert corrections \cite{ross2011dagger, kelly2019hgdagger, mandlekar2020iwr, hu2025rac}, yet relies on coarse, constant weighting that caps performance at teleoperator capability.
Alternatively, reinforcement learning learns explicit value signals, including PRMs \cite{chen2026sarm, liu2026steam, mao2026arm} and bootstrapped $Q$-functions \cite{luo2024hilserl, yang2026aloe}, to guide policy extraction through multi-step $Q$-learning \cite{wagenmaker2025dsrl, patil2026ogpo, dong2026expo} or advantage-weighted objectives \cite{peng2019advantage, frans2025cfgrl, intelligence2025pi, chen2026sarm}.

We focus on advantage-based policy extraction due to its offline stability and seamless alignment with robot foundation model post-training. 
Although recent advantage-based methods \cite{intelligence2025pi, mao2026arm, liu2026steam} incorporate human interventions, they weigh frames solely by isolated individual quality. 
This induces a learning budget imbalance in long-horizon tasks, leaving brief but difficult stages under-optimized relative to longer, simpler ones. 
In contrast, we leverage takeover statistics to estimate subtask difficulty, combining stage-level difficulty with frame progress to focus policy learning on challenging subtasks.

\begin{figure*}[!t]
\centering
\includegraphics[width=\textwidth]{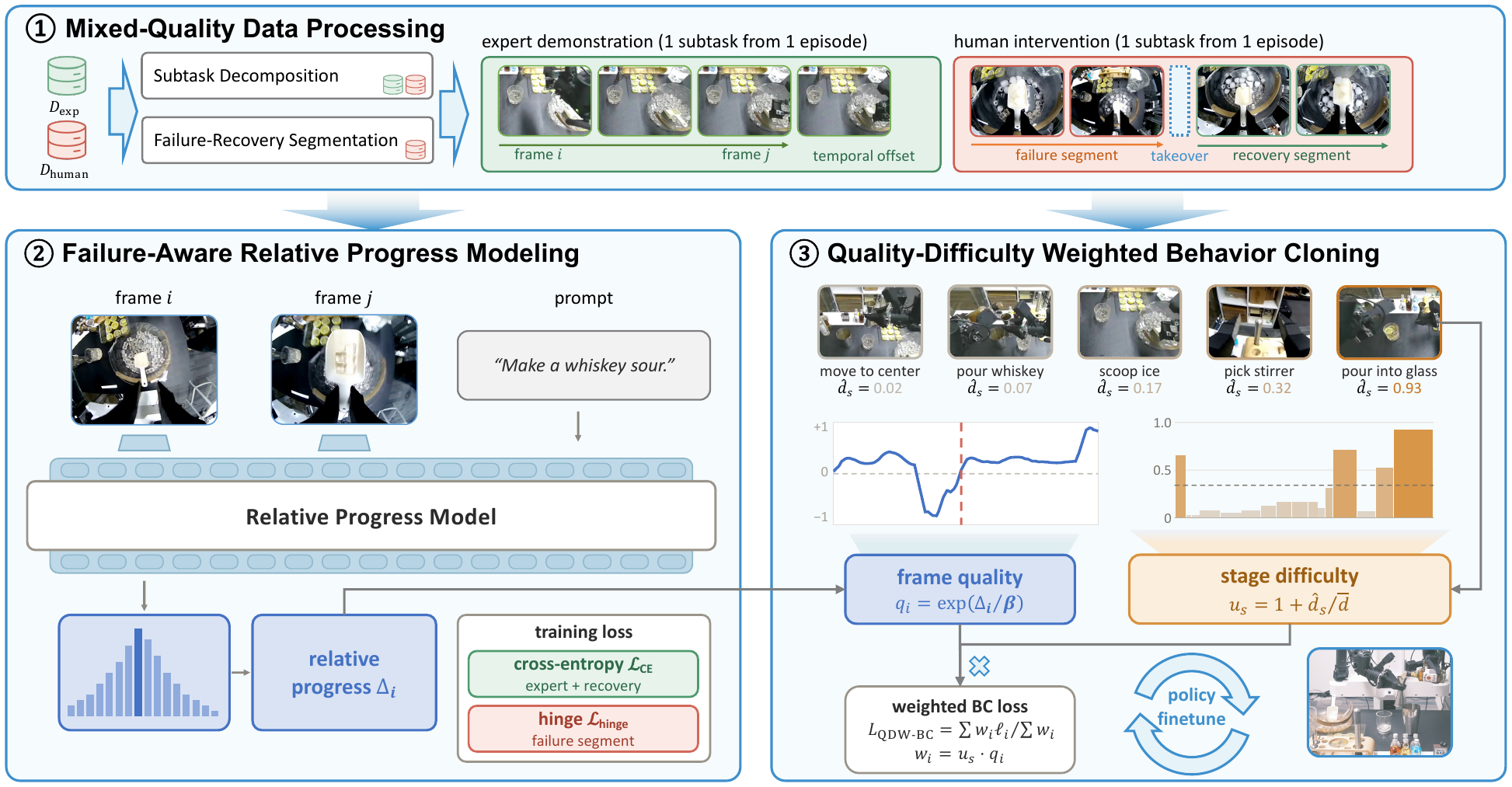}
\caption{\textbf{Methodology.}
\textbf{(1) Mixed-quality data processing.} We divide expert demonstrations
and human intervention trajectories into subtask stages. Within a stage
containing a takeover, we use the takeover timestamp to split the stage into
a pre-takeover failure segment and a post-takeover recovery segment.
\textbf{(2) Failure-aware relative progress modeling.} Given two observations
and the task instruction, \method predicts their relative progress. Expert
and recovery pairs provide exact signed temporal-offset targets, while failure
pairs receive sign-only supervision.
\textbf{(3) Quality-Difficulty Weighted Behavior Cloning.} \qdwbc weights
each training frame by its predicted relative progress and its stage's
conditional failure rate, emphasizing high-quality actions in difficult stages.}
\label{fig:framework}
\end{figure*}

\section{Methodology}
\label{sec:method}
Human intervention data inherently records operator takeovers, a single signal that our approach leverages across two distinct temporal scales. 
At the frame level, a takeover marks the transition where the policy enters an inescapable failure and human recovery begins. 
At the stage level, these same takeover timestamps measure how often each subtask fails once it is reached. 
Fig.~\ref{fig:framework} illustrates our overall post-training approach. 
Sec.~\ref{sec:method:setup} details our mixed-quality data processing, partitioning trajectories into subtask stages and intervention segments (Fig.~\ref{fig:framework}(1)). 
Sec.~\ref{sec:method:progress} presents \method, a relative progress model trained on real physical failure states (Fig.~\ref{fig:framework}(2)). 
Sec.~\ref{sec:method:qdwbc} introduces \qdwbc, a difficulty-aware policy extraction method that reweights behavior cloning by stage difficulty (Fig.~\ref{fig:framework}(3)).

\pfflush

\subsection{Mixed-Quality Data Processing}
\label{sec:method:setup}
Given expert demonstrations $\mathcal{D}_{\mathrm{exp}}$ and human intervention data $\mathcal{D}_{\mathrm{human}}$, we first partition all trajectories into subtask stages using an agent-based segmentation procedure. 
The agent is prompted with subtask names and multimodal cues indicating stage transitions, such as gripper status or object state changes, to automate segmentation without manual labeling. 
Implementation details are provided in Appendix~\ref{app:impl:data}. 
Human intervention data $\mathcal{D}_{\mathrm{human}}$ additionally records takeover timestamps. 
Within any stage containing an intervention, denoted as $s_c = \sigma(t_c)$ spanning $[\underline{s_c}, \overline{s_c})$, the takeover timestamp $t_c$ further partitions that stage into a pre-takeover failure segment $[\underline{s_c}, t_c)$ forming failure dataset $\mathcal{D}_{\mathrm{fail}}$, and a post-takeover recovery segment $[t_c, \overline{s_c})$ forming recovery dataset $\mathcal{D}_{\mathrm{rec}}$.

\subsection{Failure-aware Relative Progress Modeling}
\label{sec:method:progress}

\noindent\textbf{Relative progress model formulation.}
We construct a progress reward model parameterized by $\theta$ that quantifies relative progress toward task completion. 
Given two visual observations $o_i, o_j$ from the same episode and a natural language instruction $l$, the model predicts a categorical distribution $p_\theta(\cdot \mid o_i, o_j, l)$ over their signed temporal offset $\delta = j - i$, discretized into $B$ bins over $[-H, H]$, where $H$ denotes the maximum offset observed during training. 
We convert this predicted distribution into a scalar relative progress score $\Delta(o_i, o_j)$ via the normalized expected offset
\begin{equation}
\Delta(o_i, o_j) = \frac{1}{H} \mathbb{E}_{\delta \sim p_\theta(\cdot \mid o_i, o_j, l)} \left[\delta\right] \in [-1, 1].
\label{eq:delta}
\end{equation}
When execution advances normally, the predicted offset matches the elapsed timestep difference $j - i$. 
When progress stalls, the expected offset collapses toward $0$. 
When the policy regresses or detours, the predicted offset becomes negative. 
Unlike progress models that regress an absolute completion fraction in $[0, 1]$, \eqref{eq:delta} represents a relative quantity measuring task displacement between two frames rather than its absolute position along an episode.

\noindent\textbf{Pair construction.}
We sample observation pairs $(o_i, o_j)$ directly from the processed datasets defined in Sec.~\ref{sec:method:setup}. 
Forward pairs from expert demonstrations $\mathcal{D}_{\mathrm{exp}}$ and recovery segments $\mathcal{D}_{\mathrm{rec}}$ form positive samples with exact offset targets $\delta = j - i$, capped at $H$ because distant pairs across different subtasks render relative progress unidentifiable. 
Reversing these forward expert pairs yields synthetic negative samples $(o_j, o_i)$ with offset targets $i - j$. 
Conversely, observation pairs drawn from failure segments $\mathcal{D}_{\mathrm{fail}}$ provide genuine negative transitions driven by physical execution errors.

\noindent\textbf{Loss design.}
A natural but incorrect choice is supervising failure pairs with exact negative targets $-(j - i)$. 
As illustrated in Fig.~1 (left), real physical failures encompass stalling, detour, and physical regression. 
Stalled policies exhibit true relative offsets near $0$, while detours regress at unpredictable rates. 
Forcing these failure transitions toward fixed negative targets assigns contradictory labels to visually similar inputs, causing cross-entropy loss to corrupt value predictions across the observation space. 
Since failure transitions provide a known progress sign without a predictable magnitude, we supervise them using a one-sided hinge loss that pushes $\Delta$ below a margin $-m$ for $m > 0$. 
Stalled, drifting, and regressive transitions all satisfy this margin constraint, resolving the label contradiction. 
Denoting $\mathrm{CE}$ as the cross-entropy loss against the exact target $\delta$, the overall training objective combines both supervision forms
\begin{equation}
\begin{aligned}
&\mathcal{L}(\theta) = \mathbb{E}_{P=(o_i,o_j)} \Big[ \mathbb{1}\big[P \in \mathcal{D}_{\mathrm{fail}}\big]\max(0, \Delta(P) + m) \\
&\quad+ \mathbb{1}\big[P \in \mathcal{D}_{\mathrm{exp}} \cup \mathcal{D}_{\mathrm{rec}}\big] \cdot \mathrm{CE}\left(p_\theta(\cdot \mid P, l), \delta\right) \Big],
\end{aligned}
\label{eq:loss}
\end{equation}
where the balance between loss terms is naturally set by failure pair sampling frequency rather than an explicit hyperparameter. 
Co-sampling failure and recovery segments within the same intervention episodes eliminates domain leakage shortcuts across collection sessions.

\noindent\textbf{Lookahead clamping at takeover boundaries.}
For an episode $\tau=(o_0,\ldots,o_{T-1})$ of length $T$, let $\mathcal{T}(\tau)$ denote its set of recorded takeover timestamps.
During inference and policy extraction, we evaluate the relative progress of frame $i$ using the observation pair $(o_i, o_{i'})$ with a fixed lookahead horizon $K$. 
If a takeover timestamp $t_c$ falls within $[i, i+K)$, the pair crosses the policy-human boundary, crediting human recovery to the policy and artificially inflating progress scores. 
We eliminate this leakage by clamping the lookahead horizon at the first takeover in $\mathcal{T}(\tau)$ occurring at or after $i$, defining the per-frame progress score $\Delta_i$ as
\begin{equation}
\begin{gathered}
\Delta_i = \Delta\left(o_i, o_{i'}\right), \\
i' = \min\Big(i+K, \min\{t_c \in \mathcal{T}(\tau) \mid t_c \ge i\}, T-1\Big),
\end{gathered}
\label{eq:clamp}
\end{equation}
We similarly forbid training pairs from crossing any takeover boundary. 

\subsection{Quality-Difficulty Weighted Behavior Cloning}
\label{sec:method:qdwbc}

\noindent\textbf{Stage difficulty as a conditional failure rate.}
We define the difficulty of a task stage $s$ as the empirical conditional failure rate given that the policy reaches that stage, capturing a hazard rate rather than a raw failure count
\begin{equation}
\hat{d}_s = \frac{n^{\mathrm{fail}}_s + 1}{n^{\mathrm{reach}}_s + 2},
\label{eq:hazard}
\end{equation}
where $n^{\mathrm{fail}}_s$ denotes the number of episodes with interventions in stage $s$, satisfying $t_c \in [\underline{s}, \overline{s})$, and $n^{\mathrm{reach}}_s$ denotes the total number of episodes reaching stage $s$.
Autonomous rollout failures are incorporated into $n^{\mathrm{fail}}_s$ in the exact same manner.
Conditioning on reachability eliminates survivorship bias, ensuring that later task stages with fewer attempts are not artificially underestimated.
Because $\hat{d}_s$ reflects a property of the current policy rather than an intrinsic task attribute, it shifts dynamically as execution improves.
The smoothing in \eqref{eq:hazard} corresponds to a uniform $\mathrm{Beta}(1,1)$ prior on the failure probability, providing stable estimates when $n^{\mathrm{reach}}_s$ is small.

\noindent\textbf{Weighting frame quality by stage difficulty.}
Denoting $\ell_i$ as the behavior cloning loss on frame $i$, $N$ as the total frame count, and $S$ as the number of stages, vanilla behavior cloning decomposes across stages as
\begin{equation}
\mathcal{L}_{\mathrm{BC}} = \frac{1}{N}\sum_i \ell_i = \sum_{s=1}^{S}\rho_s\,\bar{\ell}_s.
\label{eq:bc}
\end{equation}
where $\rho_s = n_s / N$ denotes the frame fraction of stage $s$ with $n_s$ frames, and $\bar{\ell}_s = \frac{1}{n_s} \sum_{i \mid \sigma(i)=s} \ell_i$ is its average loss.
Expressed in this form, the gradient share received by stage $s$ depends solely on its frame fraction $\rho_s$, which reflects stage duration rather than difficulty.
This budget misallocation is particularly severe in long-horizon tasks, where simple subtasks dominate temporal duration and waste policy capacity.
We instead weight each frame $i$ by combining its stage difficulty and its frame-level relative progress $\Delta_i$ defined in \eqref{eq:clamp}
\begin{equation}
\begin{gathered}
\mathcal{L}_{\qdwbcm}(\pi) = \frac{1}{\sum_i w_i}\sum_i w_i\,\ell_i, \\[0.5em]
w_i = \underbrace{\Big(1 + \frac{\hat{d}_{\sigma(i)}}{\bar{d}}\Big)}_{u_{\sigma(i)} \text{ (stage difficulty)}} \cdot \underbrace{\exp\!\Big(\frac{\Delta_i}{\beta}\Big)}_{q_i \text{ (frame quality)}}, \quad \bar{d} = \sum_s \rho_s \hat{d}_s,
\end{gathered}
\label{eq:qdwbc}
\end{equation}
where $\beta > 0$ denotes a temperature hyperparameter.
Weights are rescaled to unit mean over the corpus, maintaining the loss scale and effective learning rate of vanilla BC.
The frame-level term $q_i = \exp(\Delta_i / \beta)$ discounts stalled or regressive frames, establishing quality-aware imitation \cite{peng2019advantage, chen2026sarm}.
The stage-level term $u_s = 1 + \hat{d}_s / \bar{d}$ quantifies difficulty at stage $s$, redirecting the learning budget from stage duration toward difficulty.
Crucially, $u_s$ redistributes the gradient budget across stages rather than inflating loss magnitude, as the expectation $\sum_s \rho_s u_s \equiv 2$ holds identically.

\noindent\textbf{How the weighting reallocates the learning budget.}
Regrouping \eqref{eq:qdwbc} by stage yields an objective where the stage fraction $\rho_s$ is replaced by a reweighted share $\tilde{\rho}_s$.
Assuming comparable quality factor averages across stages, the identity $\sum_s \rho_s u_s \equiv 2$ reveals this reweighted share
\begin{equation}
\tilde{\rho}_s \simeq \underbrace{\frac{\rho_s}{2}}_{\sum_s = 1/2} + \underbrace{\frac{\rho_s \hat{d}_s}{2\bar{d}}}_{\sum_s = 1/2}.
\label{eq:budget}
\end{equation}
Consequently, half of the learning budget is allocated according to stage duration, corresponding to vanilla BC, while the remaining half is allocated by the product of duration and stage difficulty.
This clean split mathematically justifies the $1 / \bar{d}$ scaling in $u_s$.
Appendix~\ref{app:derivations} provides the full derivation.

\begin{figure*}[!t]
\centering
\includegraphics[width=\textwidth]{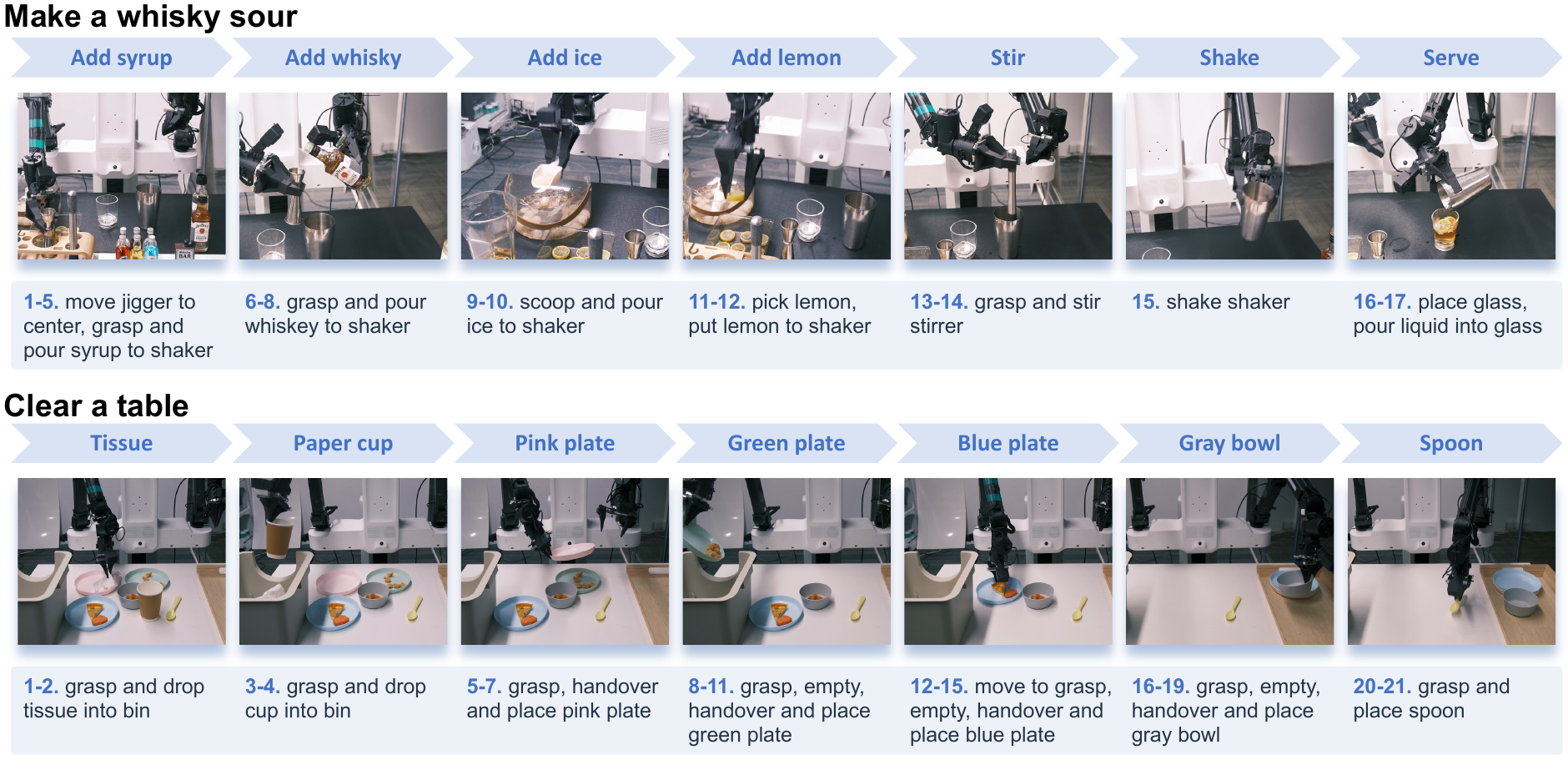}
\caption{\textbf{Subtask stages of the two real-world tasks.} 
\textsc{make a whiskey sour} (top) consists of $17$ subtask stages, while \textsc{clear a table} (bottom) comprises $21$ stages. 
Each image illustrates a group of subtasks, listed below with numbers indicating their execution order.}
\label{fig:tasks}
\par\vspace{8pt}
\includegraphics[width=\textwidth]{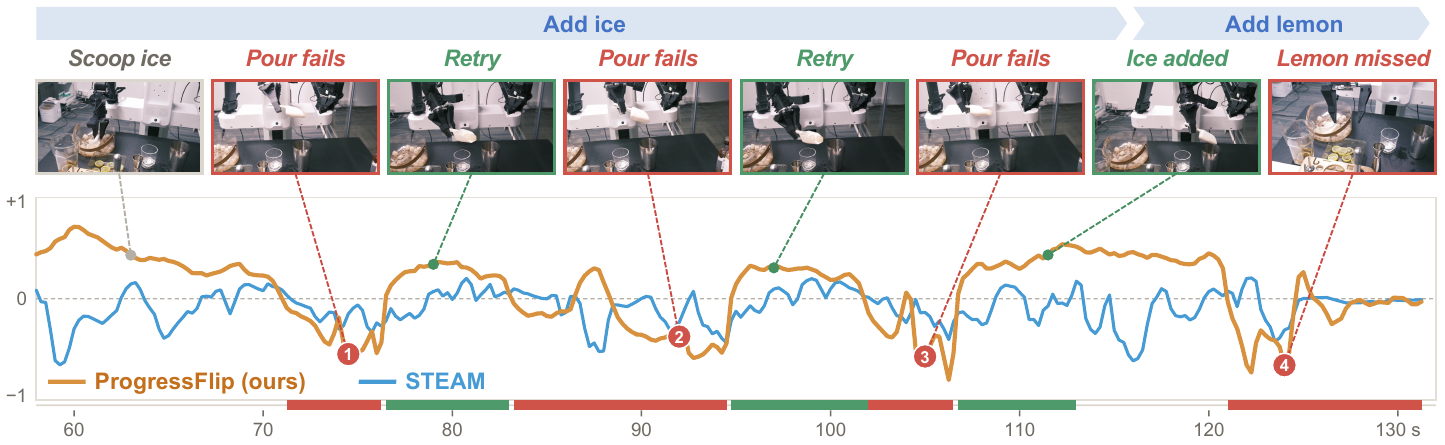}
\caption{\textbf{Relative progress visualization on an unseen \textsc{make a whiskey sour} rollout.} 
The policy experiences multiple ice-scooping failures before succeeding, followed by a missed grasp on the lemon slice. 
\method's predicted relative progress turns negative during each failure and recovers during retries, whereas STEAM misses failure onsets and recoveries. 
Red and green bars denote failure and retry segments.}
\label{fig:curves}
\end{figure*}

%
\begin{table*}[!t]
\centering
\caption{\textbf{Fidelity of progress reward predictions.} 
Evaluation across simulation and real-world platforms using two global metrics and four local failure-detection metrics. 
The held-out evaluation dataset comprises $121$ simulation and $62$ real-world intervention episodes along with $242$ simulation and $75$ real-world autonomous rollouts. 
Higher values indicate better performance.}
\label{tab:quality}
\small
\setlength{\tabcolsep}{3pt}
\sbox0{%
\begin{tabular}{@{}llcccccc@{\hspace{9pt}}cccccc@{}}
\toprule
 & & \multicolumn{6}{c}{Simulation} & \multicolumn{6}{c}{Real world} \\
\cmidrule(lr){3-8}\cmidrule(l){9-14}
 & & \multicolumn{2}{c}{Global} & \multicolumn{4}{c}{Local} & \multicolumn{2}{c}{Global} & \multicolumn{4}{c}{Local} \\
\cmidrule(lr){3-4}\cmidrule(lr){5-8}\cmidrule(lr){9-10}\cmidrule(l){11-14}
Training & PRM & \ooalign{\phantom{\textbf{100.0}}\cr\hidewidth VOC\hidewidth\cr} & CA & I-AUROC & Drop & Pre$>$Fail & Post$>$Fail & \ooalign{\phantom{\textbf{100.0}}\cr\hidewidth VOC\hidewidth\cr} & CA & I-AUROC & Drop & Pre$>$Fail & Post$>$Fail \\
\midrule
\multirow{2}{*}{Zero-shot} & RynnValue & 80.1 & 63.8 & 50.0 & 50.2 & 51.9 & 46.5 & 6.2 & 60.4 & 51.1 & 51.0 & 49.5 & 43.8 \\
 & VLAC & 53.2 & 53.5 & 50.0 & 40.5 & 51.9 & 51.5 & 74.3 & 37.7 & 48.8 & 41.1 & 51.8 & 45.1 \\
\midrule
\multirow{5}{*}{\begin{tabular}[c]{@{}l@{}}Our\\tasks\end{tabular}} & RECAP & \cellcolor{pfcell!16}99.5 & 32.1 & \cellcolor{pfcell!16}56.0 & \cellcolor{pfcell!16}52.8 & 51.4 & \cellcolor{pfcell!16}52.9 & 99.7 & 37.7 & 51.7 & 54.7 & 52.1 & 55.2 \\
 & TimeRewarder & 97.3 & \cellcolor{pfcell!16}84.3 & 51.7 & 50.0 & 51.6 & 52.0 & \cellcolor{pfcell!16}99.8 & \cellcolor{pfcell!16}79.7 & 49.2 & 48.5 & 40.7 & 41.7 \\
 & SARM & 93.8 & 73.9 & 45.7 & 47.9 & \cellcolor{pfcell!16}52.6 & 42.5 & 99.7 & 21.9 & \cellcolor{pfcell!16}56.4 & 50.6 & 56.8 & 58.3 \\
 & STEAM & \cellcolor{pfcell!45}\textbf{99.7} & 82.7 & 51.8 & 48.7 & 51.4 & 48.8 & 99.6 & 75.9 & 55.4 & \cellcolor{pfcell!16}66.7 & \cellcolor{pfcell!16}68.2 & \cellcolor{pfcell!16}60.4 \\
\cmidrule(l){2-14}
 & \method (ours) & \cellcolor{pfcell!45}\textbf{99.7} & \cellcolor{pfcell!45}\textbf{89.2} & \cellcolor{pfcell!45}\textbf{65.3} & \cellcolor{pfcell!45}\textbf{73.3} & \cellcolor{pfcell!45}\textbf{67.7} & \cellcolor{pfcell!45}\textbf{61.1} & \cellcolor{pfcell!45}\textbf{99.9} & \cellcolor{pfcell!45}\textbf{85.0} & \cellcolor{pfcell!45}\textbf{69.8} & \cellcolor{pfcell!45}\textbf{85.4} & \cellcolor{pfcell!45}\textbf{69.3} & \cellcolor{pfcell!45}\textbf{80.2} \\
\bottomrule
\end{tabular}}%
\ifdim\wd0>\textwidth\resizebox{\textwidth}{!}{\usebox0}\else\usebox0\fi
\end{table*}

\section{Experiments}
\label{sec:exp}
We evaluate our post-training approach for long-horizon tasks through three key questions.
\begin{itemize}
\item \textbf{(Q1) Progress reward modeling.} Can our progress reward model reliably estimate task progress on mixed-quality episodes, identify physical failures, and support effective policy learning (Sec.~\ref{sec:exp:value})?
\item \textbf{(Q2) Policy extraction.} Given identical progress scores, does \qdwbc improve policy performance on long-horizon tasks over existing policy extraction methods (Sec.~\ref{sec:exp:policy})?
\item \textbf{(Q3) Design choices.} How do individual components contribute to progress estimation accuracy and policy extraction (Sec.~\ref{sec:exp:ablation})?
\end{itemize}

\noindent\textbf{Baselines.}
We compare our progress reward model against SARM \cite{chen2026sarm}, designed for long-horizon progress estimation, RECAP \cite{intelligence2025pi}, and a family of methods leveraging temporal offsets or frame-pair relationships, including TimeRewarder \cite{pmlr-v306-liu26et}, STEAM \cite{liu2026steam}, VLAC \cite{zhanggeneralist}, and RynnValue \cite{huang2026rynnvalue}. 
For policy extraction, we compare \qdwbc against BC, HG-DAgger \cite{kelly2019hgdagger}, RA-BC \cite{chen2026sarm}, and AWR \cite{peng2019advantage}. 
All policy comparisons use the identical task-specific backbone and training configuration as our method. 
Appendix~\ref{app:setup:baselines} details the baseline setups.

\noindent\textbf{Tasks and datasets.}
We collect expert demonstrations and human intervention data across four RoboDojo \cite{chen2026robodojo} simulation tasks, including \textsc{stack bowls}, \textsc{put bottles into dustbin}, \textsc{pour liquid into cup}, and \textsc{classify objects}, as well as two multi-stage real-world tasks on the ARX~X5 platform, namely \textsc{make a whiskey sour} and \textsc{clear a table} (Fig.~\ref{fig:tasks}). 
Appendix~\ref{app:setup} provides detailed platform and dataset specifications. 
\pfflush

\noindent\textbf{Implementation details.}
\method builds on a SigLIP2 vision encoder and a Gemma-3-270m language encoder,
both fine-tuned end to end. All policies are fine-tuned from $\pi_0$~\cite{black2024pi_0}
except for \textsc{classify objects}, which uses $\pi_{0.5}$~\cite{intelligence2025pi_}.
Appendices~\ref{app:impl:arch} and~\ref{app:impl:hparams} detail the architecture and hyperparameters.

\subsection{Progress Reward Modeling Performance}
\label{sec:exp:value}
\pfflush

\subsubsection{Fine-grained Relative Progress Perception}
\label{sec:exp:perception}

\noindent\textbf{Protocol and metrics.}
We evaluate progress estimation independently of downstream policy learning using two global metrics (VOC, CA) and four local failure-detection metrics (I-AUROC, drop probability, Pre $>$ Fail, Post $>$ Fail) adapted from prior works \cite{qin2026retvl, liu2026steam, liu2026robovalue}. 
Global metrics assess temporal ordering and episode-level success-failure clustering. 
Local metrics evaluate sensitivity around human interventions, anchored to recorded takeover and handback timestamps without manual annotation. Appendix~\ref{app:metrics} details their computation.

\noindent\textbf{Global ordering and local failure detection.}
\method's primary strength lies in accurately detecting physical failures and recoveries. 
While all trained models achieve VOC above $93\%$ on both platforms (Table~\ref{tab:quality}), zero-shot models order frames less reliably. 
\method achieves the highest I-AUROC on held-out intervention data, outperforming STEAM by $13.5\%$ in simulation and $14.4\%$ in the real world. 
In real-world tasks, \method reaches a drop probability of $85.4\%$ and improves Post $>$ Fail to $80.2\%$ compared to STEAM's $60.4\%$. 
These results confirm that takeover-aligned supervision yields a highly responsive progress signal that drops during failures and recovers during retries. 
Furthermore, \method obtains the highest clustering accuracy (CA) on both platforms, demonstrating that its local progress estimates effectively distinguish successful from failed rollouts at the episode level.

\noindent\textbf{Qualitative comparison.}
Fig.~\ref{fig:curves} visualizes relative progress on an unseen autonomous rollout featuring repeated unassisted failures. 
\method's score drops below zero during each of the four failure events and restores to positive values upon each retry, faithfully tracking execution quality over time. 
In contrast, STEAM fluctuates but misses failure onsets and recoveries. 
TimeRewarder dips only briefly at the end of failures, while SARM exhibits delayed dips aligned with subsequent retries (Appendix~\ref{app:vis:baselines}).

\begin{table*}[!t]
\centering
\caption{\textbf{Policy extraction performance across PRMs on simulated and real-world tasks.} 
SR denotes success rate and Score represents the percentage of completed subtask stages, both reported as percentages over $50$ trials per simulation task and $30$ trials per real-world task.}
\label{tab:main}
\footnotesize
\setlength{\tabcolsep}{2pt}
\begin{tabular*}{\textwidth}{@{\extracolsep{\fill}}lcccccccccccccc@{}}
\toprule
 & \multicolumn{8}{c}{Simulation (RoboDojo)} & \multicolumn{4}{c}{Real world} & \multicolumn{2}{c}{} \\
\cmidrule(lr){2-9}\cmidrule(lr){10-13}
PRM & \multicolumn{2}{c}{\textsc{bowls}} & \multicolumn{2}{c}{\textsc{bottles}}
       & \multicolumn{2}{c}{\textsc{pour}} & \multicolumn{2}{c}{\textsc{objects}}
       & \multicolumn{2}{c}{\textsc{whiskey}} & \multicolumn{2}{c}{\textsc{table}}
       & \multicolumn{2}{c}{\emph{Avg.}} \\
\cmidrule(lr){2-3}\cmidrule(lr){4-5}\cmidrule(lr){6-7}\cmidrule(lr){8-9}\cmidrule(lr){10-11}\cmidrule(lr){12-13}\cmidrule(lr){14-15}
 & SR & Score & SR & Score & SR & Score & SR & Score
 & SR & Score & SR & Score & SR & Score \\
\midrule
 Behavior cloning & 38.0 & 43.4 & 10.0 & 28.0 & 20.0 & 20.0 & 12.0 & 25.0 & 0.0 & 40.0 & 6.7 & 18.4 & 14.4 & 29.1 \\
\midrule
 TimeRewarder & 64.0 & 68.2 & \cellcolor{pfcell!16}62.0 & \cellcolor{pfcell!16}73.9 & 46.0 & 46.0 & 42.0 & 53.1 & 23.3 & 61.8 & 43.3 & 61.4 & 46.8 & 60.7 \\
 SARM & 72.0 & 75.3 & \cellcolor{pfcell!16}62.0 & 71.1 & \cellcolor{pfcell!16}48.0 & \cellcolor{pfcell!16}48.0 & 42.0 & 55.2 & 30.0 & 73.9 & 46.7 & 61.0 & 50.1 & 64.1 \\
 STEAM & \cellcolor{pfcell!16}74.0 & \cellcolor{pfcell!16}76.7 & 60.0 & 71.4 & \cellcolor{pfcell!45}\textbf{52.0} & \cellcolor{pfcell!45}\textbf{52.0} & \cellcolor{pfcell!16}46.0 & \cellcolor{pfcell!16}59.0 & \cellcolor{pfcell!16}60.0 & \cellcolor{pfcell!16}83.9 & \cellcolor{pfcell!16}63.3 & \cellcolor{pfcell!16}75.6 & \cellcolor{pfcell!16}59.2 & \cellcolor{pfcell!16}69.8 \\
 \method (ours) & \cellcolor{pfcell!45}\textbf{80.0} & \cellcolor{pfcell!45}\textbf{82.7} & \cellcolor{pfcell!45}\textbf{72.0} & \cellcolor{pfcell!45}\textbf{79.3} & \cellcolor{pfcell!45}\textbf{52.0} & \cellcolor{pfcell!45}\textbf{52.0} & \cellcolor{pfcell!45}\textbf{54.0} & \cellcolor{pfcell!45}\textbf{66.5} & \cellcolor{pfcell!45}\textbf{76.7} & \cellcolor{pfcell!45}\textbf{90.8} & \cellcolor{pfcell!45}\textbf{90.0} & \cellcolor{pfcell!45}\textbf{93.0} & \cellcolor{pfcell!45}\textbf{70.8} & \cellcolor{pfcell!45}\textbf{77.4} \\
\bottomrule
\end{tabular*}
\end{table*}

\begin{table}[!t]
\centering
\caption{\textbf{Policy extraction performance on real-world tasks.} RA-BC, AWR, and \qdwbc use identical \method progress scores.}
\label{tab:policy-extraction}
\small
\setlength{\tabcolsep}{3pt}
\begin{tabular}{@{}lcccccc@{}}
\toprule
 & \multicolumn{2}{c}{\textsc{whiskey}} & \multicolumn{2}{c}{\textsc{table}} & \multicolumn{2}{c}{\emph{Avg.}} \\
\cmidrule(lr){2-3}\cmidrule(lr){4-5}\cmidrule(l){6-7}
Method & SR & Score & SR & Score & SR & Score \\
\midrule
 Behavior cloning & 0.0 & 40.0 & 6.7 & 18.4 & 3.3 & 29.2 \\
 HG-DAgger & 30.0 & 64.7 & 46.7 & 64.8 & 38.3 & 64.8 \\
 AWR & 26.7 & 60.4 & 60.0 & \cellcolor{pfcell!16}87.9 & 43.3 & 74.2 \\
 RA-BC & \cellcolor{pfcell!16}46.7 & \cellcolor{pfcell!16}78.0 & \cellcolor{pfcell!16}66.7 & 80.6 & \cellcolor{pfcell!16}56.7 & \cellcolor{pfcell!16}79.3 \\
 \qdwbc (ours) & \cellcolor{pfcell!45}\textbf{76.7} & \cellcolor{pfcell!45}\textbf{90.8} & \cellcolor{pfcell!45}\textbf{90.0} & \cellcolor{pfcell!45}\textbf{93.0} & \cellcolor{pfcell!45}\textbf{83.3} & \cellcolor{pfcell!45}\textbf{91.9} \\
\bottomrule
\end{tabular}
\end{table}

\subsubsection{Progress-weighted Policy Extraction}
\label{sec:exp:vwpe}
\pfflush
We evaluate whether our predicted relative progress provides effective supervision for downstream policy extraction. 
Holding the policy backbone and training dataset fixed within each task, we compare four PRMs under the identical \qdwbc objective (Table~\ref{tab:main}). 

Across all six tasks, \method achieves an average success rate of $70.8\%$, outperforming the strongest baseline (STEAM) by $11.6\%$ in success rate and $7.6\%$ in stage score. 
In simulation, \method leads by a significant margin on three tasks and ties STEAM on \textsc{pour liquid into cup}. 
This performance pattern occurs because \textsc{stack bowls}, \textsc{put bottles into dustbin}, and \textsc{classify objects} contain explicit policy failure entry trajectories during intervention collection. 
In contrast, failures in \textsc{pour liquid into cup} stem from bottle misalignment that spills liquid onto the table, representing an irreversible physical failure. 
Consequently, human operators must intervene preemptively during data collection before liquid spills, making it challenging to capture distinct policy failure entry segments.
In real-world deployment, \method demonstrates substantial gains over all baselines on complex long-horizon tasks. 
On \textsc{make a whiskey sour}, \method achieves $76.7\%$ success, outperforming STEAM's $60.0\%$. 
On \textsc{clear a table}, it reaches $90.0\%$ success, surpassing STEAM's $63.3\%$. 
These substantial improvements highlight that accurately detecting physical failure transitions and human recoveries is essential for guiding policy capacity toward critical bottlenecks in long-horizon real-world manipulation. 
Combined with our local progress perception metrics, these results confirm that failure-sensitive relative progress estimation provides superior supervisory signals for policy learning.

\subsection{Policy Performance Analysis}
\label{sec:exp:policy}
We evaluate five policy extraction methods on two multi-stage real-world tasks over $30$ evaluation trials per task (Table~\ref{tab:policy-extraction}). 
Vanilla behavior cloning fails on both tasks with a $3.3\%$ average success rate, confirming that expert demonstrations alone are insufficient for complex, long-horizon manipulation. 
Incorporating human-in-the-loop interventions through HG-DAgger enables policies to learn from interactive experience, raising average success to $38.3\%$. 
Methods leveraging Progress Reward Models overall surpass HG-DAgger, demonstrating the importance of fine-grained credit assignment on mixed-quality data. 

Among progress-weighted methods utilizing identical \method scores, \qdwbc achieves the highest performance, reaching $76.7\%$ success on \textsc{make a whiskey sour} and $90.0\%$ on \textsc{clear a table}. 
By modulating frame-level progress with stage-wise failure rates, \qdwbc outperforms RA-BC by an average of $26.6\%$ in success rate. 
This substantial margin confirms that rebalancing the learning budget by stage difficulty successfully concentrates policy capacity on brief, high-difficulty bottlenecks, whose task-specific distributions are detailed in Appendix Fig.~\ref{fig:corpora}.

\subsection{Ablation Studies}
\label{sec:exp:ablation}

\begin{table}[!t]
\centering
\caption{\textbf{Ablation studies of \method components on real-world tasks.} All variants are evaluated on failure windows immediately preceding takeovers. Higher values indicate better failure perception. Hyphens denote non-convergence.}
\label{tab:ablation}
\small
\setlength{\tabcolsep}{3.5pt}
\resizebox{\pfnarrow}{!}{%
\begin{tabular}{@{}lcccc@{}}
\toprule
 & \multicolumn{2}{c}{\textsc{whiskey}} & \multicolumn{2}{c}{\textsc{table}} \\
\cmidrule(lr){2-3}\cmidrule(lr){4-5}
Variant & I-AUROC $\uparrow$ & Drop Prob. $\uparrow$ & I-AUROC $\uparrow$ & Drop Prob. $\uparrow$ \\
\midrule
\method (ours)        & \textbf{81.3} & \textbf{99.4} & \textbf{91.8} & \textbf{100.0} \\
\midrule
w/o Stage segmentation     & 61.4 & 48.5 & -- & -- \\
Exact failure offsets      & 77.9 & 93.0 & 60.1 & 54.0 \\
w/o Lookahead clamping     & 65.8 & 91.2 & 78.0 & 89.7 \\
\bottomrule
\end{tabular}}
\end{table}
\pfflush

We analyze the contribution of each core component in \method by ablating one component at a time on both real-world tasks (Table~\ref{tab:ablation}). 
To isolate the effect of lookahead clamping, all variants use evaluation windows immediately preceding takeovers. 
Metric values are therefore comparable internally within this ablation study.

\noindent\textbf{Stage segmentation.}
Removing stage segmentation degrades I-AUROC from $81.3$ to $61.4$ and drop probability from $99.4\%$ to $48.5\%$ on \textsc{make a whiskey sour}, while causing training divergence on \textsc{clear a table}. 
This confirms that pairing frames across the entire episode dilutes local failure supervision, validating subtask stage decomposition as a crucial design for consistent progress estimation.

\noindent\textbf{Sign-based failure supervision.}
Replacing sign-based hinge loss with exact-offset cross-entropy degrades performance across both tasks, causing a $31.7\%$ drop in I-AUROC and a $46.0\%$ drop in drop probability on \textsc{clear a table}. 
Because human takeovers signify failure entry without prescribing an exact regression magnitude, forcing fixed temporal targets introduces contradictory labels. 
Sign-based supervision accommodates this inherent ambiguity without imposing ungrounded temporal targets.

\noindent\textbf{Lookahead clamping.}
Allowing evaluation frame pairs to cross takeover boundaries reduces I-AUROC by $15.5\%$ on \textsc{make a whiskey sour} and $13.8\%$ on \textsc{clear a table}. 
On \textsc{make a whiskey sour}, the mean progress score during failure windows reverses from a negative $-0.16$ to a positive $+0.15$, confirming that unclamped pairs falsely credit human recovery to autonomous execution. 
Clamping eliminates this progress leakage.

Together, these results validate stage decomposition, sign-based supervision, and lookahead clamping as indispensable components for reliable physical failure detection.

\section{Conclusions}
\label{sec:conclusion}
In this work, we presented a post-training approach for robot foundation models on mixed-quality interaction data. 
By leveraging human interventions as free annotations of progress sign flips, \method captures real physical failures via sign-only supervision while learning exact temporal offsets on expert and recovery transitions. 
To resolve stage-level budget imbalances, we introduced \qdwbc to reweight policy updates using both frame-level progress and stage-wise conditional failure rates. 
Across four RoboDojo simulation tasks and two multi-stage real-world tasks, our approach substantially outperforms state-of-the-art baselines. 
Overall, our approach demonstrates that human intervention structure serves as an effective supervisory signal for scalable, long-horizon robot learning.

\pfrestorefloats
\bibliographystyle{unsrtnat}
\bibliography{refs}

\clearpage
\beginappendix
\renewcommand{\twocolumn}[1][]{\par\noindent #1\par}
\makeatletter
\renewenvironment{figure*}[1][]{\@float{figure}[!htbp]}{\end@float}
\renewenvironment{table*}[1][]{\@float{table}[!htbp]}{\end@float}
\makeatother
%

\section{Robot Setup, Tasks and Baselines}
\label{app:setup}
This section describes the robot platforms, tasks, datasets, and baselines used in our experiments.

\subsection{Real-World Platform and Tasks}
\label{app:setup:real}

\noindent\textbf{Hardware.}
Our tabletop platform comprises two ARX~X5 arms mounted side by side on a fixed torso. 
Three Orbbec RGB cameras observe the workspace, including one on each wrist and one mounted on the head. 
The head camera provides an overall workspace view, while the wrist cameras capture fine-grained interactions between grippers and objects. 
Episodes are recorded at $20$\,fps, containing three synchronized image streams and the robot state.

\begin{figure}[htbp]
\centering
\includegraphics[width=\pfphoto]{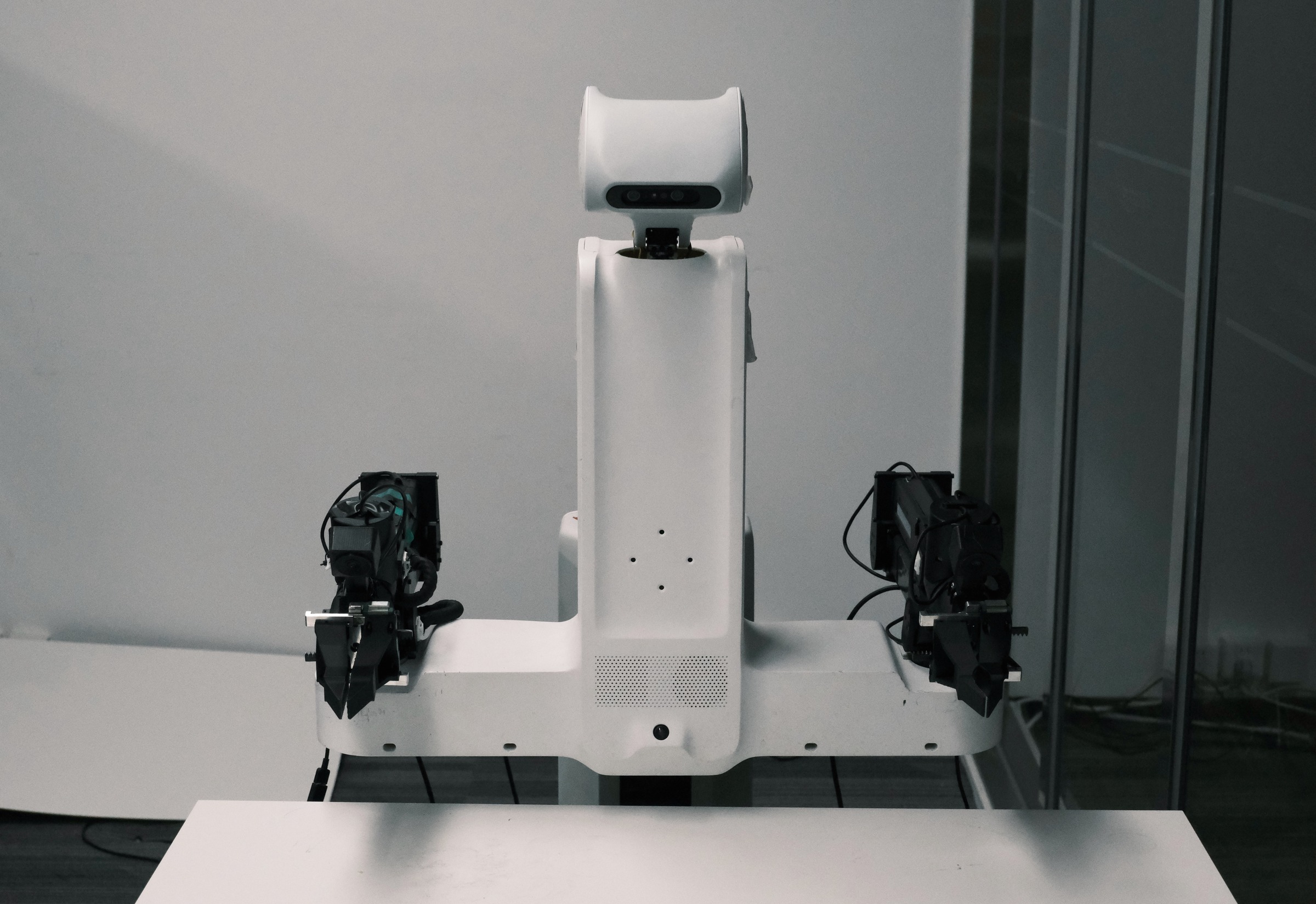}
\caption{\textbf{Real-robot platform.} 
Two ARX~X5 arms equipped with grippers perform bimanual tabletop manipulation. 
Three Orbbec cameras mounted on the head and wrists provide an overview of the workspace and close-up views of object interactions.}
\label{fig:robot_setup}
\end{figure}

\begin{figure*}[!t]
\centering
\includegraphics[width=\textwidth]{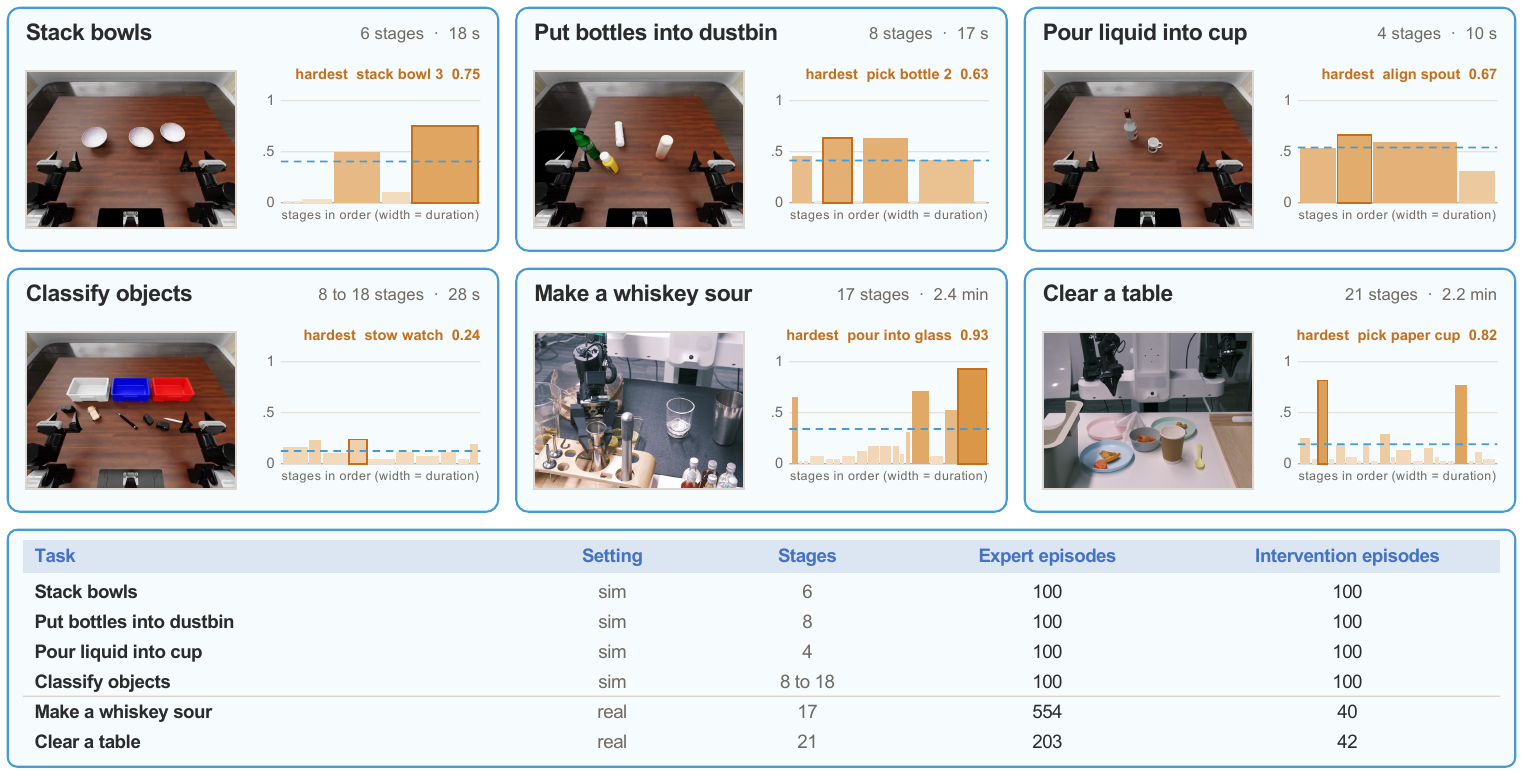}
\caption{\textbf{Stage difficulty distributions and training dataset statistics.} 
Each panel illustrates the simulated layout or a third-person view of the real-world setup. 
Bar width indicates the temporal duration fraction of each subtask stage, shading denotes the empirical conditional failure rate, and dashed lines mark the frame-weighted mean difficulty. 
Most tasks concentrate difficulty within a few critical stages. 
The embedded table lists the number of expert demonstrations and human intervention episodes used for policy training.}
\label{fig:corpora}
\end{figure*}

\noindent\textbf{Teleoperation and intervention collection.}
We collect expert demonstrations using a leader-follower interface where an operator moves a pair of kinematically matched leader arms while the robot reproduces the motion. 
The same interface enables human interventions during policy execution, allowing the operator to take control during imminent failures and return control to the policy upon recovery. 
The system logs control modes and intervention timestamps alongside visual and proprioceptive observations. 
Consequently, an intervention episode contains both autonomous execution and human-guided recovery, potentially involving multiple takeover events. 
Episode counts and takeover counts reflect distinct quantities, as illustrated in Fig.~\ref{fig:corpora}.

\begin{itemize}
\item \textbf{\textsc{make a whiskey sour}} comprises $17$ subtask stages and lasts approximately $2.4$ minutes per demonstration. The robot grasps and positions a jigger, measures syrup and liquor into a shaker, adds ice and lemon, stirs and shakes the mixture, and pours the drink into a serving glass, combining tool use with coordinated bimanual manipulation of bottles, small utensils, and containers. Task success requires completing the entire sequence through the final pour.
\item \textbf{\textsc{clear a table}} comprises $21$ subtask stages and lasts approximately $2.2$ minutes per demonstration. The robot disposes of a tissue and a paper cup, clears and stacks three plates and a bowl, and places a spoon at its target location. The right arm grasps each container, empties residual contents when necessary, and hands it to the left arm for placement, requiring four arm-to-arm handovers. Task success requires disposing of or clearing all designated objects.
\end{itemize}

\subsection{Simulation Benchmark}
\label{app:setup:sim}
We evaluate four RoboDojo \cite{chen2026robodojo} tasks using a simulated ARX~X5 bimanual robot. 
Each observation consists of three camera streams and a $14$-dimensional joint state vector, recorded at $25$\,fps. 
Fig.~\ref{fig:corpora} illustrates representative task layouts.

\begin{itemize}
\item \textbf{\textsc{stack bowls}} comprises six subtask stages of grasping and stacking, requiring precise spatial alignment between bowls.
\item \textbf{\textsc{put bottles into dustbin}} comprises eight subtask stages involving bottle grasping, bimanual arm-to-arm handover, and disposal.
\item \textbf{\textsc{pour liquid into cup}} comprises a four-stage sequence where the robot grasps a container, aligns it with the target cup, and pours.
\item \textbf{\textsc{classify objects}} requires sorting four to nine objects into designated containers, where each object contributes a pick stage and a place stage, yielding $8$ to $18$ stages per episode.
\end{itemize}

Each task provides $100$ expert demonstrations and $100$ human intervention episodes for policy training, where an intervention episode may contain multiple takeover events. 
The first three tasks feature fixed stage sequences, whereas \textsc{classify objects} varies in sequence length depending on the number of objects. 
For stage-difficulty estimation, we group stages by object class in \textsc{classify objects} and merge the two grasping modes in \textsc{put bottles into dustbin}. 
Appendix~\ref{app:impl:data} details the automated stage decomposition procedure.

\subsection{Baselines}
\label{app:setup:baselines}

\subsubsection{Progress Reward Models}

\noindent\textbf{SARM.} \cite{chen2026sarm} SARM estimates long-horizon task progress by predicting the active subtask stage and within-stage progress. It incorporates subtask annotations and dataset-level stage-duration priors for calibration. We evaluate its progress differences over the same temporal window as \method.

\noindent\textbf{TimeRewarder.} \cite{pmlr-v306-liu26et} TimeRewarder learns normalized signed temporal distances between frame pairs from expert demonstrations to construct per-step rewards. We accumulate these rewards over the identical temporal window to obtain per-frame scores.

\noindent\textbf{STEAM.} \cite{liu2026steam} STEAM learns normalized temporal offsets from expert observation pairs using reversed pairs as synthetic negative supervision. It converts predicted offset distributions into advantage scores. We evaluate STEAM on observation pairs spanning the same temporal window as \method.

\noindent\textbf{RECAP.} \cite{intelligence2025pi} RECAP learns a distributional value function from task-normalized Monte Carlo returns with per-step costs and a terminal failure penalty. We compute differences in its expected values over the same temporal window as per-frame progress scores. Its training corpus incorporates autonomous rollout data, and its evaluation results are reported in Table~\ref{tab:quality}.

\noindent\textbf{RynnValue.} \cite{huang2026rynnvalue} RynnValue predicts remaining task execution time from an episode prefix. We evaluate its released weights zero-shot and negate the predicted remaining time to represent progress. Each query uses eight head-camera frames sampled uniformly from the prefix at $0.5$-second intervals in simulation and $1.0$-second intervals in the real world.

\noindent\textbf{VLAC.} \cite{zhanggeneralist} VLAC predicts progress increments between frame pairs. We apply its released weights zero-shot to head-camera frame pairs sampled one second apart, accumulating predictions into progress curves following its official inference procedure. For both zero-shot baselines, progress curves are interpolated to every frame and converted into local scores using a $32$-frame difference.

\noindent\textbf{Inference cost.}
Table~\ref{tab:prm_cost} compares the inference cost of all Progress Reward Models on a single NVIDIA A800 GPU for \textsc{make a whiskey sour} recorded at $20$\,fps. We measure model forward-pass time exclusively, excluding video decoding and CPU preprocessing overhead. Per-query latency is measured with a batch size of one to simulate online deployment. GPU time per minute of video reflects the batched offline labeling settings and query frequencies in Table~\ref{tab:quality}. \method issues one pair query per frame and incurs one third of STEAM's compute cost because STEAM runs a three-model ensemble. TimeRewarder and SARM require less compute because they encode single head-camera views using CLIP, whereas zero-shot models query once per second but rely on large vision-language models.

\begin{table}[htbp]
\centering
\caption{\textbf{Computational inference cost of Progress Reward Models.} Measured on a single NVIDIA A800 GPU for \textsc{make a whiskey sour} at $20$\,fps. Images per query denote camera views multiplied by frames, where STEAM repeats queries across three ensemble members. GPU time is reported per minute of video under batched offline labeling settings.}
\label{tab:prm_cost}
\small
\setlength{\tabcolsep}{3pt}
\sbox0{%
\begin{tabular}{@{}lccccc@{}}
\toprule
PRM & Params & Images & Queries & Latency & GPU time \\
 & (B) & / query & / s & (ms / query) & (s / min) \\
\midrule
RynnValue & 9.57 & 8 & 1 & 177.7 & 8.4 \\
VLAC & 8.08 & 2 & 1 & 157.8 & 6.9 \\
RECAP & 0.73 & 3 & 20 & 45.7 & 12.4 \\
TimeRewarder & 0.15 & 2 & 20 & 3.8 & 0.6 \\
SARM & 0.27 & 1 & 20 & 8.5 & 1.7 \\
STEAM & 4.22 & 6 $\times$ 3 & 20 & 105.2 & 37.9 \\
\method (ours) & 1.41 & 6 & 20 & 35.1 & 12.6 \\
\bottomrule
\end{tabular}}%
\ifdim\wd0>\pfnarrow\resizebox{\pfnarrow}{!}{\usebox0}\else\usebox0\fi   
\end{table}

\subsubsection{Policy Extraction Strategies}

\noindent\textbf{Behavior cloning (BC).} BC trains exclusively on expert demonstrations with uniform weights without utilizing progress reward predictions. It serves as the unweighted expert-only baseline across policy evaluations.

\noindent\textbf{HG-DAgger.} \cite{kelly2019hgdagger} HG-DAgger augments expert demonstrations with human intervention data and trains with uniform sample weights without score-based reweighting.

\noindent\textbf{RA-BC.} \cite{chen2026sarm} RA-BC combines reward-based sample filtering and reweighting for policy learning. We apply identical \method progress scores as in AWR and \qdwbc to determine its per-frame training weights.

\noindent\textbf{AWR.} \cite{peng2019advantage} AWR weights training samples exponentially using predicted progress scores. Our implementation retains the frame-quality weighting term in \eqref{eq:qdwbc} while omitting stage-difficulty weighting, isolating the specific contribution of stage-wise difficulty scaling.

\section{Implementation Details}
\label{app:impl}
This section details stage decomposition, model architecture, and the training and inference settings of \method and \qdwbc. Table~\ref{tab:hparams} summarizes the hyperparameters.

\subsection{Automated Stage Decomposition}
\label{app:impl:data}
We segment every episode into subtask stages using an agent-based decomposition procedure that converts prompt instructions into rule-based segmentation scripts. 
For each task, the prompt provides the task name, ordered subtask sequence, and heuristic cues indicating subtask completion. 
These cues primarily fall into two types. 
The first type includes gripper state transitions, such as closing to grasp or opening to release an object. 
The second type involves observable object state changes, such as a container resting on its target, a plate being emptied, or a bottle returning upright after pouring. 
The agent grounds gripper cues directly in robot state channels, applying hysteresis thresholds and minimum duration filters to suppress noise. 
For object-state cues, it utilizes proprioceptive proxies, such as wrist position or roll recovery, or evaluates candidate frames from camera views when proprioception carries no object state information. 
Failed retries remain within the active stage, extending from the initial attempt to successful completion. 
Coordinated bimanual events distinguish handovers from direct placements. 
Scripts undergo structural validation requiring strictly increasing contiguous boundaries, while rare unresolvable episodes receive manual verification. 
Final boundaries and metadata are stored alongside takeover records.

\noindent\textbf{Example.} 
For \textsc{make a whiskey sour}, the prompt defines $17$ subtasks with corresponding completion cues. 
Most cues map to gripper state changes, where the agent anchors boundaries to five right-arm and three left-arm object excursions in fixed sequence across expert episodes. 
Subtasks without gripper events rely on spatial proxies, such as wrist position thresholds for jigger alignment or scoop trajectory points for ice retrieval. 
These rules segment all $100$ expert demonstrations automatically. 
For intervention data where takeovers split single grasps into multiple attempts, the agent matches gripper excursions to prototype object grasp positions and aligns them using monotonic dynamic programming. 
This procedure automatically segments $34$ out of $40$ intervention episodes, leaving six complex cases for manual verification. 
Table~\ref{tab:stages} summarizes stage counts and end-of-stage heuristics across all six tasks.

\begin{table*}[!t]
\centering
\caption{\textbf{Automated stage decomposition.} Stage counts, subtask sequences, and the heuristics that end each stage. Unless noted otherwise, grasping stages end at gripper closure and placing stages end at gripper opening, and repeated attempts remain in the current stage.}
\label{tab:stages}
\small
\renewcommand{\arraystretch}{1.15}
\begin{tabular}{@{}p{0.15\textwidth}p{0.05\textwidth}p{0.27\textwidth}p{0.44\textwidth}@{}}
\toprule
\textbf{Task} & \textbf{Stages} & \textbf{Subtask sequence} & \textbf{End-of-stage heuristics} \\
\midrule
\textsc{make a whiskey sour} & $17$ & Grasp and position the jigger, add syrup and liquor to the shaker, add ice and lemon, stir, shake, place the glass, and pour & Five right-gripper and three left-gripper object excursions in a fixed order, the right wrist reaching a reference position ends positioning and transfer stages, the scoop leaving the ice bucket ends ice scooping, and the right gripper opening toward the glass ends shaking \\
\addlinespace
\textsc{clear a table} & $21$ & Dispose of a tissue and a paper cup, grasp, empty, hand over, and place three plates and a bowl, pick and place a spoon & Each stage ends at a gripper closure, right for grasping and disposal and left for placement, a left closure must fall inside each handover stage, and emptying ends when the right wrist last crosses back past $20\%$ of the path from the dustbin to the table center \\
\addlinespace
\textsc{stack bowls} & $6$ & Grasp and place each of three bowls & The number of separate objects on the table, read from the head image at each release, separates stacking from moving the first bowl or a failed stack, the next grasp target separates a retry from a new bowl, and episodes starting with a bowl at the center skip the first two stages \\
\addlinespace
\textsc{pour liquid into cup} & $4$ & Grasp the bottle, align the spout with the cup, pour, and place the bottle & The working arm holds its gripper closed for more than $40\%$ of the episode, alignment ends when the wrist roll first exceeds $60\%$ of its peak deviation, and pouring ends when the roll falls back below $80\%$ of the peak \\
\addlinespace
\textsc{put bottles into dustbin} & $8$ & Pick up and dispose of each of four bottles & Exactly four left-gripper closure and opening cycles per episode, since every bottle enters the dustbin from the left arm, and a right-gripper grasp enclosing a left closure marks a handover \\
\addlinespace
\textsc{classify objects} & $8$ to $18$ & Pick up and place each of four to nine objects & The end-effector position at release identifies the target basket, a release at the table center followed by a closure of the other arm marks a transfer, and a release on the table followed by a closure of the same arm marks a dropped object and a retry \\
\bottomrule
\end{tabular}
\end{table*}

\subsection{Model Architecture}
\label{app:impl:arch}

\noindent\textbf{Visual and language encoders.}
A shared SigLIP2-so400m-patch14-224 encoder processes camera images at $224\times224$ resolution. 
A linear-GELU-dropout projector maps pooled visual embeddings to $512$ dimensions, and we average embeddings across all three camera views to form the observation representation. 
Gemma-3-270m encodes natural language task instructions, truncated to a maximum of $200$ tokens. 
We mean-pool its final hidden states over valid tokens and project the result to $512$ dimensions.

\noindent\textbf{Relative progress head.}
We concatenate two observation embeddings in temporal sequence with the instruction embedding, yielding a $1536$-dimensional vector. 
Layer normalization followed by a linear-GELU-dropout-linear MLP head produces $B$ offset logits over discrete signed strides $\{-B/2, \ldots, -1, 1, \ldots, B/2\}$. 
The normalized expected stride yields the relative progress score in \eqref{eq:delta}, preserving observation order and outputting an estimate bounded within $[-1, 1]$.

\noindent\textbf{Policy backbone.}
We fine-tune $\pi_0$ \cite{black2024pi_0} across all tasks except \textsc{classify objects}, where we fine-tune $\pi_{0.5}$ \cite{intelligence2025pi_}. 
Because initial $\pi_0$ checkpoints achieve zero success on \textsc{classify objects}, we adopt the stronger $\pi_{0.5}$ starting backbone for this task. 
All methods within a task evaluate on identical backbones and action representations.

\subsection{Training and Inference Settings}
\label{app:impl:hparams}
Table~\ref{tab:hparams} details the architectural parameters, optimization schedules, inference settings, and policy training hyperparameters. 
We end-to-end fine-tune both encoders and the prediction head using AdamW in bfloat16 precision, applying gradient checkpointing to the language model. 
Expert and recovery transitions are trained using cross-entropy with label smoothing, whereas failure transitions are trained using the one-sided hinge loss. 
Their sampling ratio dynamically balances the two loss terms.

Our default progress model configures $B=32$ offset bins and a maximum offset of $H=64$ frames, whereas \textsc{make a whiskey sour} configures $B=16$ and $H=128$. 
During inference, all tasks evaluate with a lookahead horizon of $K=32$ frames under takeover-boundary clamping. 
We modulate these frame-quality scores with stage-wise difficulty estimates to train \qdwbc, assigning unsegmented frames a neutral difficulty factor $u=2$ and normalizing weights to unit mean per dataset prior to policy learning.

\begin{table*}[!t]
\centering
\caption{\textbf{Training and inference hyperparameters.} Settings for \method and \qdwbc. Task-specific exceptions are indicated explicitly.}
\label{tab:hparams}
\small
\begin{tabular}{@{}ll@{}}
\toprule
\textbf{Parameter (symbol)} & \textbf{Value} \\
\midrule
\multicolumn{2}{@{}l}{\emph{Relative progress model} (Sec.~\ref{sec:method:progress})} \\
\quad Vision encoder                          & SigLIP2-so400m-patch14-224 \\
\quad Language model                          & Gemma-3-270m \\
\quad Fusion hidden dimension                 & 512 \\
\quad Frames per pair                         & 2 \\
\quad Max prompt length                       & 200 tokens \\
\quad Camera views                            & 3 (head, left wrist, right wrist) \\
\quad Image resolution                        & $224\times224$ \\
\quad Encoders fine-tuned                     & both (vision and language) \\
\quad Number of offset bins ($B$)             & 32 (16 on \textsc{make a whiskey sour}) \\
\quad Max training offset ($H$)               & 64 frames (128 on \textsc{make a whiskey sour}) \\
\quad Hinge margin ($m$)                      & 0.1 \\
\quad Label smoothing                         & 0.05 \\
\quad Dropout                                 & 0.1 \\
\quad Optimizer                               & AdamW ($\beta_1{=}0.9$, $\beta_2{=}0.95$, $\epsilon{=}10^{-8}$) \\
\quad Learning rate / schedule                & $5\times10^{-5}$ / constant \\
\quad Warm-up                                 & 100 steps (simulation), 500 steps (real robot) \\
\quad Weight decay                            & $10^{-5}$ \\
\quad Gradient clipping                       & 10.0 \\
\quad Precision                               & bfloat16 \\
\quad Batch size (global)                     & 512 \\
\quad Training steps                          & 8{,}000 (simulation and \textsc{make a whiskey sour}), 16{,}000 (\textsc{clear a table}) \\
\quad Intervention sampling weight            & per task, set so failure pairs are $1.0$ to $1.5\%$ of a batch \\
\midrule
\multicolumn{2}{@{}l}{\emph{Relative progress inference} (Eq.~\ref{eq:clamp})} \\
\quad Lookahead ($K$)                         & 32 frames \\
\quad Takeover clamping                       & enabled \\
\quad Batch size                              & 512 \\
\midrule
\multicolumn{2}{@{}l}{\emph{Weighted behavior cloning} (Sec.~\ref{sec:method:qdwbc})} \\
\quad Stage hazard ($\hat d_s$)                    & $(n^{\mathrm{fail}}_s{+}1)/(n^{\mathrm{reach}}_s{+}2)$, Laplace-smoothed \\
\quad Stage difficulty ($u_s$)                & $1+\hat d_s/\bar{d}$, where $\bar{d}=\sum_s \rho_s\hat d_s$ \\
\quad Neutral difficulty (unlabelled frames)  & 2 \\
\quad Frame quality ($q_i$)                   & $\exp(\Delta_i/\beta)$, clipped at $c$ (never active since $\Delta\in[-1,1]$) \\
\quad AWR temperature ($\beta$)               & 1.0 \\
\quad Quality clip ($c$)                      & 20 \\
\quad Weight normalisation                    & per-dataset, each rescaled to mean $1$ \\
\quad Policy backbone                         & $\pi_0$ ($\pi_{0.5}$ for \textsc{classify objects}) \\
\quad Action chunk size                       & 50 frames / $2.0$\,s (simulation), 20 frames / $1.0$\,s (real robot) \\
\quad Action dimension                        & 14 (simulation), 28 (real robot) \\
\quad Optimizer                               & AdamW ($\beta_1{=}0.9$, $\beta_2{=}0.95$, $\epsilon{=}10^{-8}$) \\
\quad Learning rate / schedule                & $2.5\times10^{-5}$ peak / cosine decay to $2.5\times10^{-6}$ \\
\quad Warm-up / decay horizon                 & 1{,}000 / 30{,}000 steps \\
\quad EMA decay                               & 0.99 \\
\quad Gradient clipping                       & 1.0 \\
\quad Batch size (global)                     & 128 \\
\quad Training steps                          & 10{,}000 in simulation, 30{,}000 on real robot \\
\bottomrule
\end{tabular}
\end{table*}

\begin{figure*}[!t]
\centering
\includegraphics[width=0.86\textwidth]{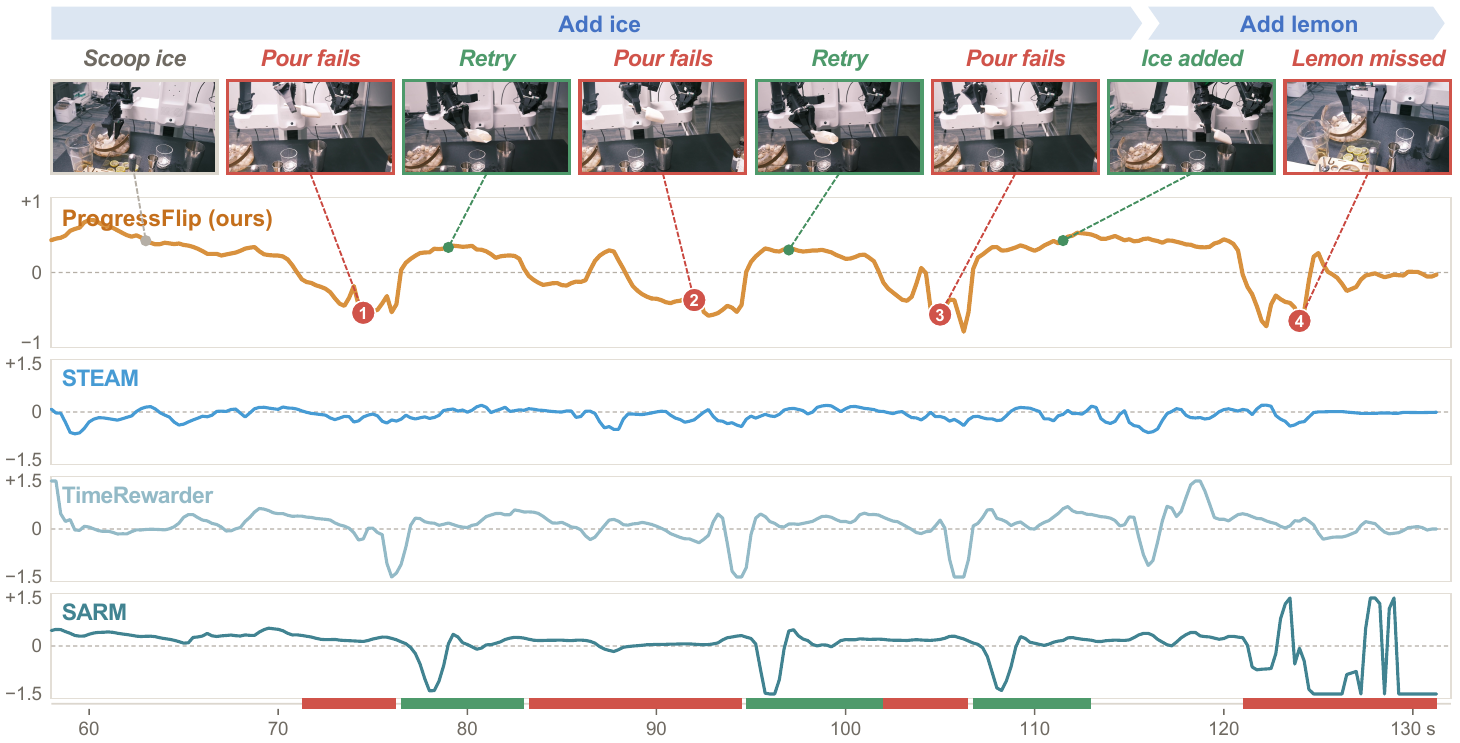}
\caption{\textbf{All PRMs on the failed rollout of Fig.~\ref{fig:curves}.} STEAM fluctuates throughout
without a consistent drop at failure onset or rise at recovery. TimeRewarder dips briefly near the end of each failed ice pour and
spikes before the lemon grasp. SARM dips at the start of each retry rather than
during the failure, and oscillates between the clipping limits while the lemon
grasp fails.}\label{fig:vis:baselines}
\par\vspace{1.0em}
\includegraphics[width=0.86\textwidth]{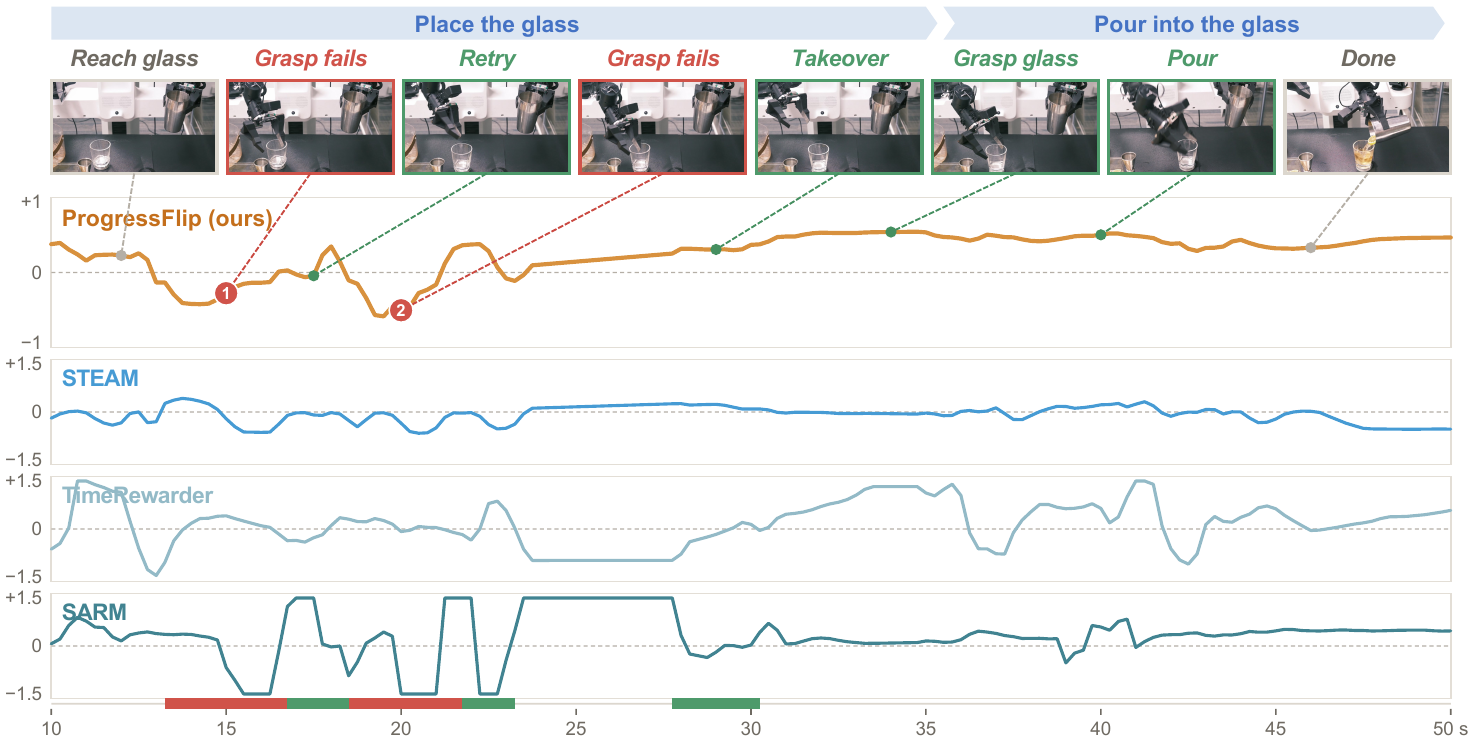}
\caption{\textbf{All PRMs on a rollout with retries at the glass.} The policy fails twice to grasp the serving glass, an operator briefly takes over, and the policy then grasps the glass and pours. \method's score turns negative during both failed grasps and stays positive after the takeover. STEAM fluctuates without a consistent response to either failure. TimeRewarder drops sharply at the first failed grasp but also during the pause before the takeover and during the pour. SARM jumps between the clipping limits across both failed grasps and the pause that follows.}\label{fig:vis:baselines2}
\end{figure*}

\section{Relative Progress Results Visualization}\label{app:vis}
This section provides per-frame relative progress visualizations. 
Appendix~\ref{app:vis:baselines} compares all Progress Reward Models on two \textsc{make a whiskey sour} rollouts, while Appendix~\ref{app:vis:tasks} illustrates \method on three representative episodes per task. 
Reference frames appear above each curve, with dashed lines anchoring them to specific timesteps. 
Colored strips beneath the time axes mark failure segments in red, recovery or retry segments in green, and unrecovered failure stretches in gray. 
For intervention episodes, a red segment highlights up to six seconds of the active stage prior to a recorded takeover, followed by a green segment denoting human-controlled recovery. 
For expert demonstrations, green segments indicate retried grasping or placing operations.

\subsection{Comparison of PRMs}\label{app:vis:baselines}
Both rollouts are held out and recorded in unseen environments. 
The first rollout is a fully autonomous episode with unassisted failures, whereas the second rollout is a human intervention episode where repeated policy failures when grasping the glass prompt an operator to take over control and complete the task. 

\twocolumn[{%
\subsection{ProgressFlip on Simulation and Real-world Tasks}\label{app:vis:tasks}
Each figure illustrates an expert demonstration, a human intervention episode, and a failed autonomous rollout of one task.
\par\vspace{1.2em}
\begin{minipage}{\textwidth}\centering
\includegraphics[width=0.95\textwidth]{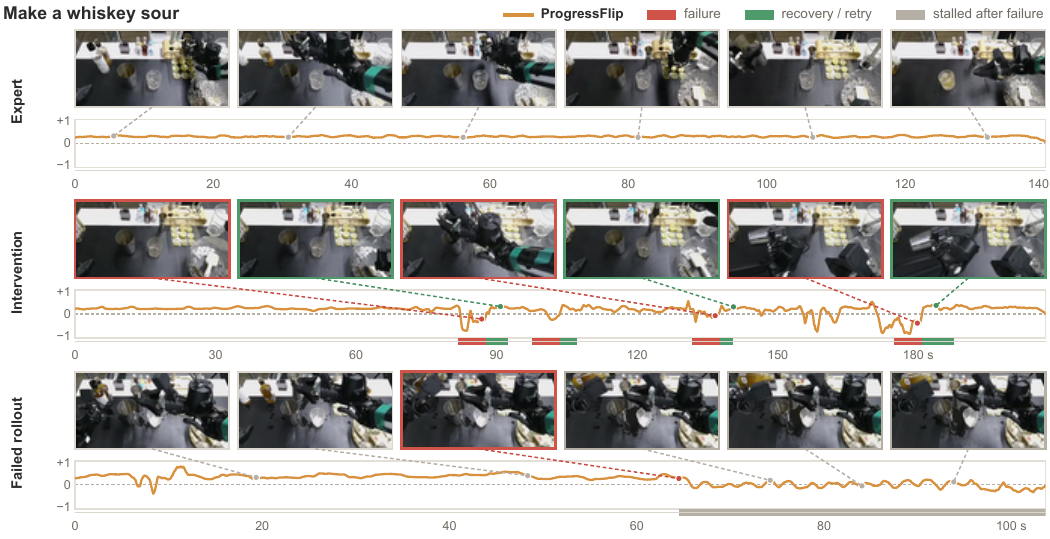}
\captionof{figure}{\textbf{\textsc{make a whiskey sour}.} The expert demonstration maintains a steady positive score across all $17$ subtask stages. In the intervention episode, predicted relative progress drops sharply prior to each takeover and restores to positive values upon human recovery. The failed autonomous rollout stalls at approximately $64$\,s, after which the progress score oscillates near zero until episode termination.}\label{fig:vis:whiskey}
\end{minipage}
\par\vspace{1.2em}
\begin{minipage}{\textwidth}\centering
\includegraphics[width=0.95\textwidth]{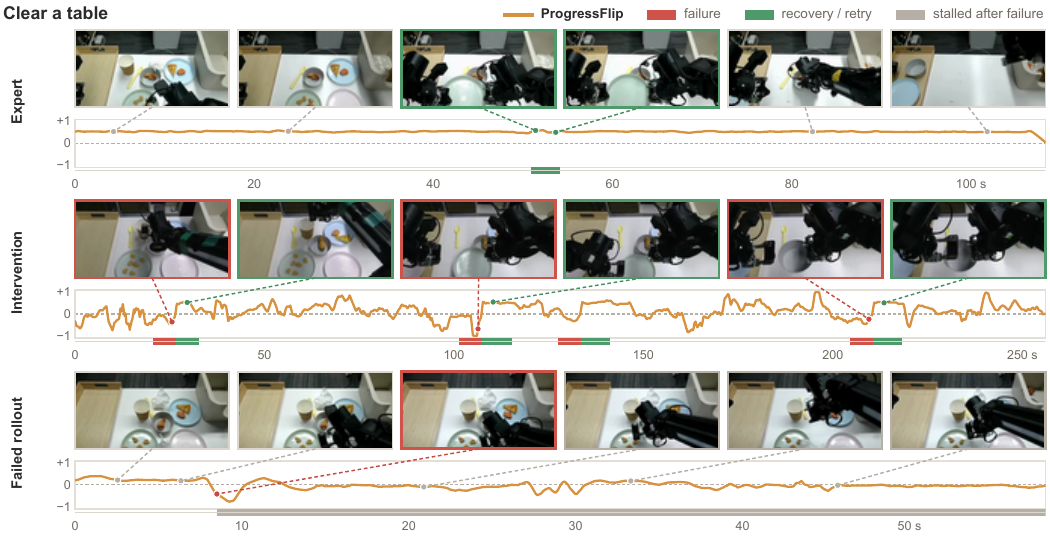}
\captionof{figure}{\textbf{\textsc{clear a table}.} The expert demonstration maintains progress near $+0.5$, exhibiting a shallow dip during a retried plate handover. The intervention episode contains four takeovers, each preceded by a progress drop and followed by a recovery to expert levels. The failed rollout experiences an early failure within the first $9$\,s and remains near zero for the remaining $50$\,s.}\label{fig:vis:table}
\end{minipage}
\vspace{0.5em}}]

\twocolumn[{%
\begin{minipage}{\textwidth}\centering
\includegraphics[width=0.95\textwidth]{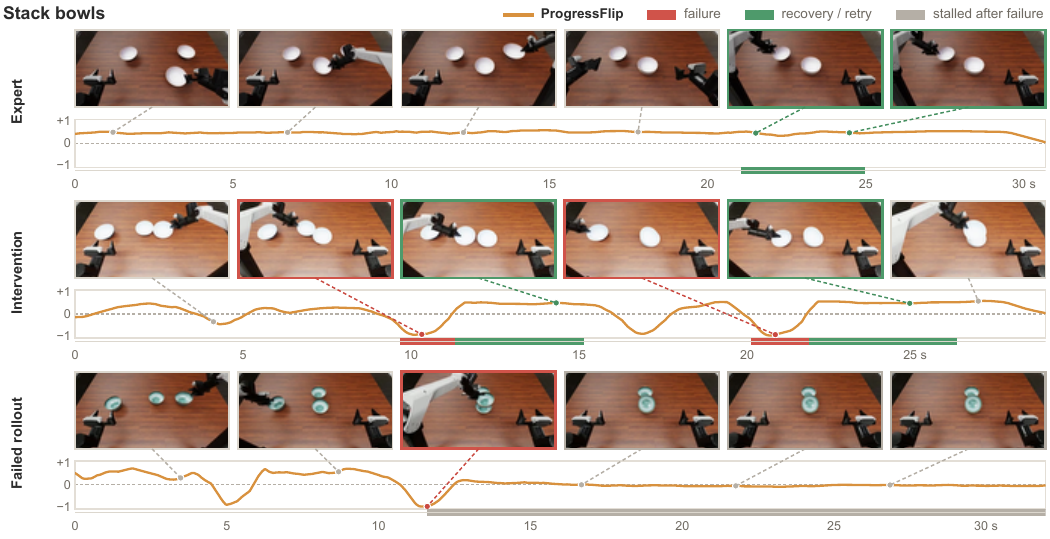}
\captionof{figure}{\textbf{\textsc{stack bowls}.} A regrasp during the expert demonstration causes a minor score decrease without turning negative. Both takeovers in the intervention episode occur after progress drops to $-1$, with scores returning to expert levels under human control. Following a failed stacking attempt at $12$\,s, the autonomous rollout score remains at zero for the remaining $20$\,s.}\label{fig:vis:bowls}
\end{minipage}
\par\vspace{1.2em}
\begin{minipage}{\textwidth}\centering
\includegraphics[width=0.95\textwidth]{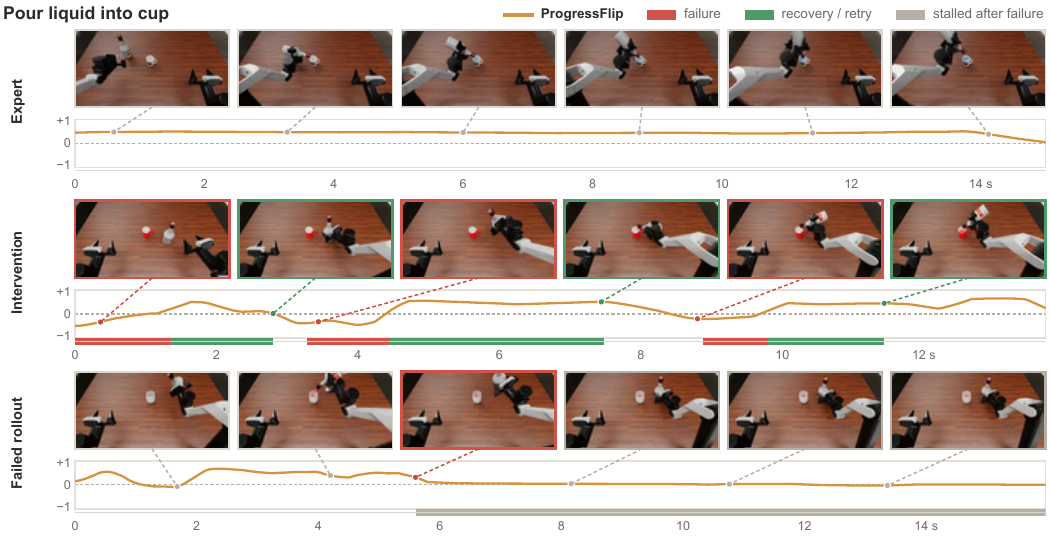}
\captionof{figure}{\textbf{\textsc{pour liquid into cup}.} Due to short episode durations of approximately $15$\,s, the three takeovers in the intervention episode occur in close succession, each preceded by a negative progress segment. The failed rollout ceases progress at $6$\,s, maintaining a flat zero score thereafter.}\label{fig:vis:pour}
\end{minipage}
\vspace{0.5em}}]

\twocolumn[{%
\begin{minipage}{\textwidth}\centering
\includegraphics[width=0.95\textwidth]{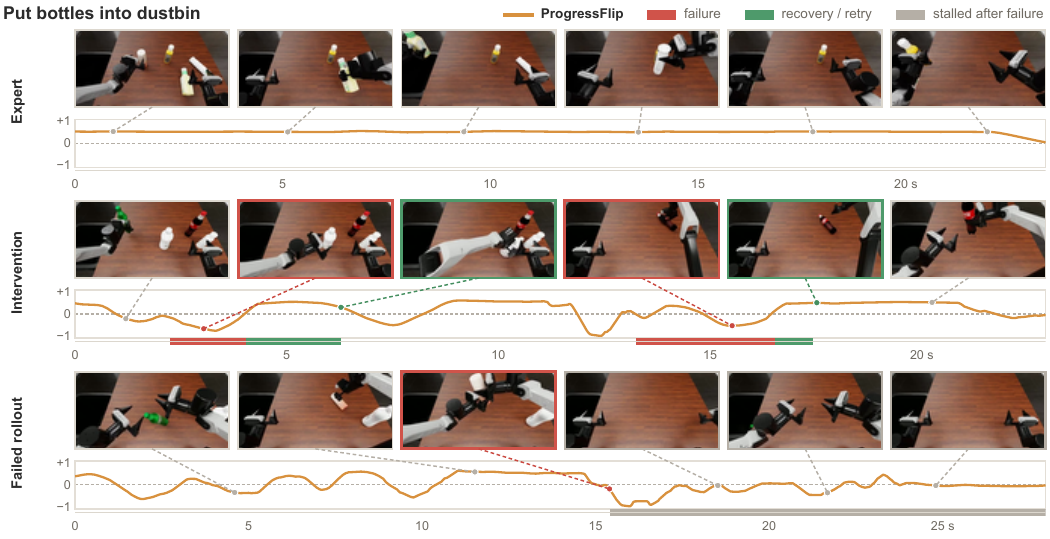}
\captionof{figure}{\textbf{\textsc{put bottles into dustbin}.} The expert demonstration maintains positive progress near $+0.5$ throughout. Both takeovers in the intervention episode follow a negative progress segment. The failed rollout exhibits a sharp score drop at $15$\,s and remains at or below zero afterwards.}\label{fig:vis:bottles}
\end{minipage}
\par\vspace{1.2em}
\begin{minipage}{\textwidth}\centering
\includegraphics[width=0.95\textwidth]{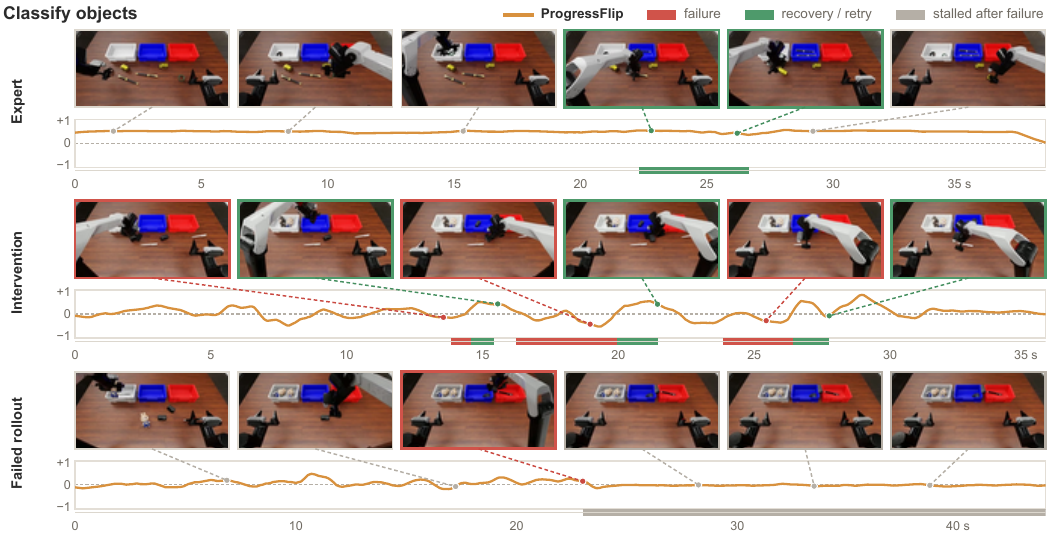}
\captionof{figure}{\textbf{\textsc{classify objects}.} A retried grasp in the expert demonstration produces a shallow progress dip. In the intervention episode, predicted progress drops prior to each of the three takeovers and recovers following human intervention. The failed rollout stalls at $23$\,s, maintaining a flat zero score thereafter.}\label{fig:vis:cls}
\end{minipage}
\vspace{0.5em}}]

\twocolumn[{%
\section{Attention Visualization}\label{app:attention}
\vspace{0.3em}
\begin{minipage}{\textwidth}\centering
\includegraphics[width=\textwidth]{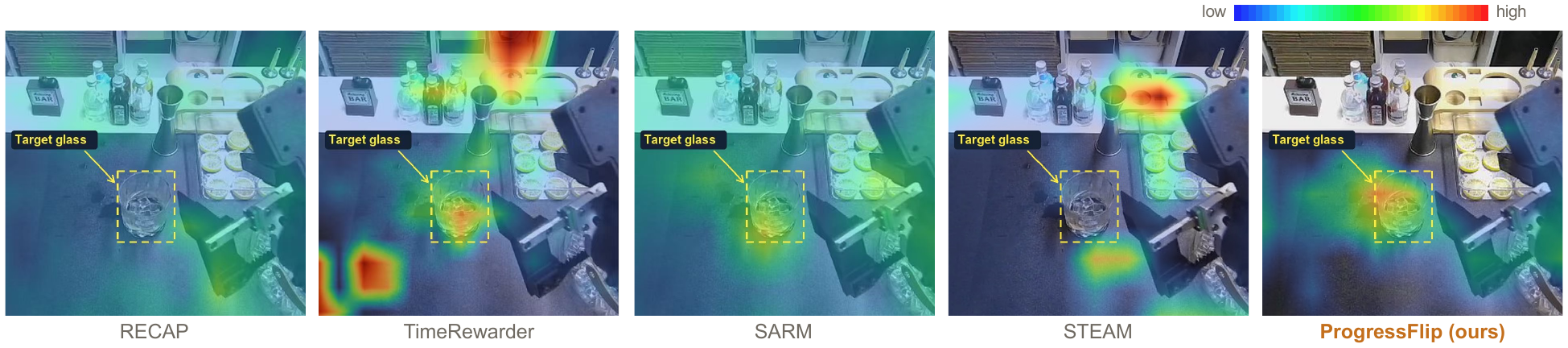}
\captionof{figure}{\textbf{Spatial attention distribution across Progress Reward Models.} All panels illustrate the identical head-camera frame. 
The outlined glass denotes the pouring target, where warmer colors indicate higher attention weights.}\label{fig:attn}
\end{minipage}
\vspace{0.5em}}]

\subsection{Comparison Setup}
Fig.~\ref{fig:attn} compares spatial attention maps on the identical head-camera frame during the final pouring stage of \textsc{make a whiskey sour}. 
We extract raw attention weights in evaluation mode and display the spatial map for the active camera frame. 
The visualization utilizes frame pair $(t-128, t)$ for \method and STEAM, $(t-1, t)$ for TimeRewarder, the single active frame for RECAP, and history ending at timestep $t$ for SARM. 
These input configurations differ from the standard $32$-frame lookahead window utilized during policy weighting.

\subsection{Model-Specific Extraction}

\noindent\textbf{ProgressFlip.}
We average SigLIP2 pooling-query to patch attention weights across all heads to obtain a $16\times16$ grid. 
Removing padding-associated rows yields a $12\times16$ spatial map of the visible camera frame. 
These visual weights are extracted prior to temporal and language feature fusion.

\noindent\textbf{STEAM.}
We apply the identical pooling attention extraction and padding removal as in \method, retaining the spatial map from the ensemble member selected by the conservative minimum aggregation rule.

\noindent\textbf{TimeRewarder.}
We average final-layer CLIP ViT-B/16 CLS-to-patch attention weights across all heads, yielding a $14\times14$ grid. 
Images are resized to a short side of $256$ pixels and center-cropped to $224\times224$, with attention mapped back exclusively to the visible crop.

\noindent\textbf{RECAP.}
We average final critic layer CLS-to-visual-token attention weights across heads. 
The three camera token groups form separate $16\times16$ maps, where we retain the head-camera map and remove padding-associated rows. 
Language and non-visual tokens are excluded.

\noindent\textbf{SARM.}
We average final-layer CLS-to-patch attention weights from its frozen CLIP ViT-B/32 encoder, producing a $7\times7$ spatial map for the active image. 
Image preprocessing resizes the short side to $224$ pixels followed by a $224\times224$ center crop. 
This visual encoder map is extracted prior to history, state, and language feature processing in the stage predictor.

\subsection{Rendering and Interpretation}
We normalize each spatial map according to its visible image area, utilizing a shared display scale fixed at the pooled $99.5$th percentile. 
All maps undergo the identical contrast transformation and Turbo colormap mapping, bilinearly resized to their corresponding image regions and overlaid on a common crop surrounding the right arm and target glass. 
The glass outline serves as a visual annotation only. 
\method exhibits a highly localized response concentrated around the target glass, whereas baseline models emphasize uninformative background regions or display more diffuse attention patterns. 
Because model architectures and feature extraction layers differ across baselines, these spatial visualizations serve qualitative comparison purposes rather than definitive causal claims.

\section{Derivations}
\label{app:derivations}

This section derives the learning-budget split in \eqref{eq:budget}, using the notation of Sec.~\ref{sec:method:qdwbc}.
\label{app:deriv:weight}

In practice, the weights $w_i$ in \eqref{eq:qdwbc} are rescaled to unit mean within each source dataset. 
We write the rescaled weight as $\bar w_i = w_i / Z_i$, where $Z_i$ is the mean weight of the dataset containing frame $i$. 
Since $Z_i$ is constant within each dataset, we absorb it into the frame-quality term by defining $\tilde q_i = q_i / Z_i$, so that $\bar w_i = u_{\sigma(i)}\,\tilde q_i$. 
For each stage $s$, let
\begin{equation}
Q_s = \sum_{i \mid \sigma(i)=s} \tilde q_i ,
\qquad
\bar{\ell}^{\,q}_s = \frac{1}{Q_s}\sum_{i \mid \sigma(i)=s} \tilde q_i\,\ell_i
\end{equation}
denote the total quality mass of stage $s$ and its quality-weighted average loss. 
All frames in stage $s$ share the same factor $u_s$, so grouping both sums in \eqref{eq:qdwbc} by stage gives
\begin{equation}
\begin{aligned}
\mathcal{L}_{\qdwbcm}
&= \frac{\sum_{s} u_s \sum_{i \mid \sigma(i)=s} \tilde q_i\,\ell_i}{\sum_{s'} u_{s'} Q_{s'}}
= \sum_{s=1}^{S} \tilde{\rho}_s\,\bar{\ell}^{\,q}_s , \\
\tilde{\rho}_s &= \frac{u_s Q_s}{\sum_{s'} u_{s'} Q_{s'}} .
\end{aligned}
\label{eq:app_regroup}
\end{equation}
This objective has the same form as \eqref{eq:bc}, with the stage fraction $\rho_s$ replaced by the reweighted share $\tilde{\rho}_s$, and the shares again sum to one.

If the average quality factor is comparable across stages, then $Q_s \simeq \bar q\, n_s$ with $\bar q = \frac{1}{N}\sum_i \tilde q_i$. 
Substituting this into \eqref{eq:app_regroup} and dividing the numerator and denominator by $\bar q N$ yields
\begin{equation}
\tilde{\rho}_s \simeq \frac{u_s \rho_s}{\sum_{s'} u_{s'} \rho_{s'}} .
\label{eq:app_share}
\end{equation}
By the definition of $\bar d$ in \eqref{eq:qdwbc}, the denominator equals
\begin{equation}
\begin{aligned}
\sum_s \rho_s u_s
&= \sum_s \rho_s\Big(1+\frac{\hat{d}_s}{\bar{d}}\Big) \\
&= \sum_s \rho_s + \frac{1}{\bar{d}}\sum_s \rho_s \hat{d}_s
= 1 + 1 = 2 ,
\end{aligned}
\label{eq:app_const}
\end{equation}
which holds identically and does not depend on the data.

Substituting \eqref{eq:app_const} and $u_s = 1 + \hat{d}_s/\bar{d}$ into \eqref{eq:app_share} gives
\begin{equation}
\tilde{\rho}_s \simeq \frac{\rho_s}{2}\Big(1+\frac{\hat{d}_s}{\bar{d}}\Big)
= \frac{\rho_s}{2} + \frac{\rho_s \hat{d}_s}{2\bar{d}} ,
\end{equation}
which is \eqref{eq:budget}. 
Summing over stages, the first term contributes $\sum_s \rho_s/2 = 1/2$ and the second contributes $\sum_s \rho_s \hat{d}_s/(2\bar{d}) = \bar{d}/(2\bar{d}) = 1/2$. 
Half of the learning budget therefore follows stage duration as in vanilla BC, and the other half follows the product of stage duration and stage difficulty.

\section{Progress Evaluation Metrics}
\label{app:metrics}
This section defines the metrics in Table~\ref{tab:quality}, following ReTVL~\cite{qin2026retvl} and STEAM~\cite{liu2026steam}. 
All metrics are computed on frame-level scores, so models with different output scales can be compared.

\noindent\textbf{Value-order correlation (VOC).}
VOC measures whether progress accumulates with time along successful expert episodes. For each episode we sample $200$ frame pairs $(i,j)$ with $i<j$, sum the frame-level scores between the two frames, and compute the Spearman rank correlation between these sums and the gaps $j-i$. Scores are averaged over episodes within a task and then over tasks.

\noindent\textbf{Clustering accuracy (CA).}
CA measures whether the score separates successful from failed autonomous rollouts without a threshold tuned on the labels. We average the frame-level score over each episode, fit a two-component Gaussian mixture to these averages, and label episodes in the higher-mean component as successful. CA is the mean of the per-class recalls against the recorded outcomes, since the rollout corpora are imbalanced. We average rather than sum the score, because a summed score mainly reflects episode length when failed rollouts run to the time limit.

\noindent\textbf{Local evaluation windows.}
The local metrics are anchored to the recorded takeover time $t_{\mathrm{int}}$ and the time $t_{\mathrm{end}}$ at which control returns to the policy. Four windows of width $W$ bracket each event, with $W$ equal to one second in simulation and two seconds on the real robot,
\begin{equation}
\begin{aligned}
\mathcal{F}^{\mathrm{pre}}  &= [\,t_{\mathrm{int}}-K-3W,\;\; t_{\mathrm{int}}-K-2W\,) ,\\[0.2em]
\mathcal{F}^{\mathrm{fail}} &= [\,t_{\mathrm{int}}-K-W,\;\; t_{\mathrm{int}}-K\,) ,\\[0.2em]
\mathcal{F}^{\mathrm{post}} &= [\,t_{\mathrm{end}}+W,\;\; t_{\mathrm{end}}+2W\,) ,\\[0.2em]
\mathcal{F}^{\mathrm{late}} &= [\,t_{\mathrm{int}}+5W,\;\; t_{\mathrm{int}}+6W\,) ,
\end{aligned}
\label{eq:windows}
\end{equation}
where the offset by the lookahead $K$ in \eqref{eq:clamp} keeps the failure window from reaching across the takeover. The policy is still on track in $\mathcal{F}^{\mathrm{pre}}$, about to fail in $\mathcal{F}^{\mathrm{fail}}$, and back on track in $\mathcal{F}^{\mathrm{post}}$, while $\mathcal{F}^{\mathrm{late}}$ lies at a fixed offset after the takeover regardless of how long the operator stays in control. A faithful score should dip over the failure window and recover on either side of it. We report takeover events that admit all four windows.

\noindent\textbf{Intervention AUROC (I-AUROC).}
I-AUROC is the probability that a frame after recovery is scored above a frame before the takeover,
\begin{equation}
\text{I-AUROC} = \Pr_{f_i\in\mathcal{F}^{\mathrm{late}},\, f_j\in\mathcal{F}^{\mathrm{fail}}}\big[\,\Delta(f_i) > \Delta(f_j)\,\big] .
\label{eq:iauroc}
\end{equation}

\noindent\textbf{Drop probability.}
Drop probability is the fraction of takeover events whose mean score over $\mathcal{F}^{\mathrm{fail}}$ falls below the median score of the same episode.

\noindent\textbf{Pre $>$ Fail and Post $>$ Fail.}
These metrics are the fractions of takeover events for which the mean score in $\mathcal{F}^{\mathrm{pre}}$, respectively $\mathcal{F}^{\mathrm{post}}$, exceeds the mean score in $\mathcal{F}^{\mathrm{fail}}$. Pre $>$ Fail tests whether the score drops as the policy drifts toward failure, and Post $>$ Fail tests whether it recovers once the policy resumes control. For these event-level metrics an exact tie between the two means counts as one half, matching the rank convention of I-AUROC.

\end{document}